\documentclass[twoside]{article}
\usepackage[T1]{fontenc}
\usepackage{amsmath}
\usepackage{amssymb}
\usepackage[accepted]{aistats2025}
\usepackage{graphicx}
\usepackage{bm}
\usepackage[colorlinks=true,linkcolor=blue,citecolor=blue,urlcolor=blue]{hyperref}

\usepackage[round]{natbib}

\begin{document}

%

%

\twocolumn[

\aistatstitle{ Hypernym Bias: Unraveling Deep Classifier Training Dynamics through the Lens of Class Hierarchy}



\aistatsauthor{Roman Malashin \And Valeria Yachnaya \And Alexander Mullin}

\aistatsaddress{%
  Saint-Petersburg State University of Aerospace Instrumentation \\
  Pavlov Institute of Physiology, RAS
} ]

\begin{abstract}
  We investigate the training dynamics of deep classifiers by examining how
  hierarchical relationships between classes evolve during training. Through
  extensive experiments, we argue that the learning process in classification
  problems can be understood through the lens of label clustering. Specifically,
  we observe that networks tend to distinguish higher-level (hypernym)
  categories in the early stages of training, and learn more specific (hyponym)
  categories later. We introduce a novel framework to track the evolution of the
  feature manifold during training, revealing how the hierarchy of class
  relations emerges and refines across the network layers. Our analysis
  demonstrates that the learned representations closely align with the semantic
  structure of the dataset, providing a quantitative description of the
  clustering process. Notably, we show that in the hypernym label space, certain
  properties of neural collapse appear earlier than in the hyponym label space,
  helping to bridge the gap between the initial and terminal phases of learning.
  We believe our findings offer new insights into the mechanisms driving
  hierarchical learning in deep networks, paving the way for future advancements
  in understanding deep learning dynamics.
\end{abstract}

\section{INTRODUCTION}
Deep neural networks have demonstrated remarkable capabilities in learning
complex functions, yet the underlying reasons for their success remain an open
question. In this work, we aim to shed light on one crucial aspect of this
puzzle: \textbf{What patterns do neural networks tend to learn first, and which ones do
they learn last?}
The question has been studied for a long time, but valuable interpretations
continue to emerge. One important contribution is the concept of simplicity bias
proposed by \cite{arpit2017closer}, which suggests that neural networks tend to
learn simple (frequent) patterns followed by more complex or noisy patterns.
Neural collapse is another phenomenon that characterizes the terminal phase of
classifier training \citep{papyan2020prevalence}. During this phase variance of
penultimate features within the same class converges to zero, while the class
means are arranged in the feature space according to Equiangular Tight
Frame(ETF) structure.

\begin{figure*}[h]
  \centering
  \includegraphics[width=15cm]{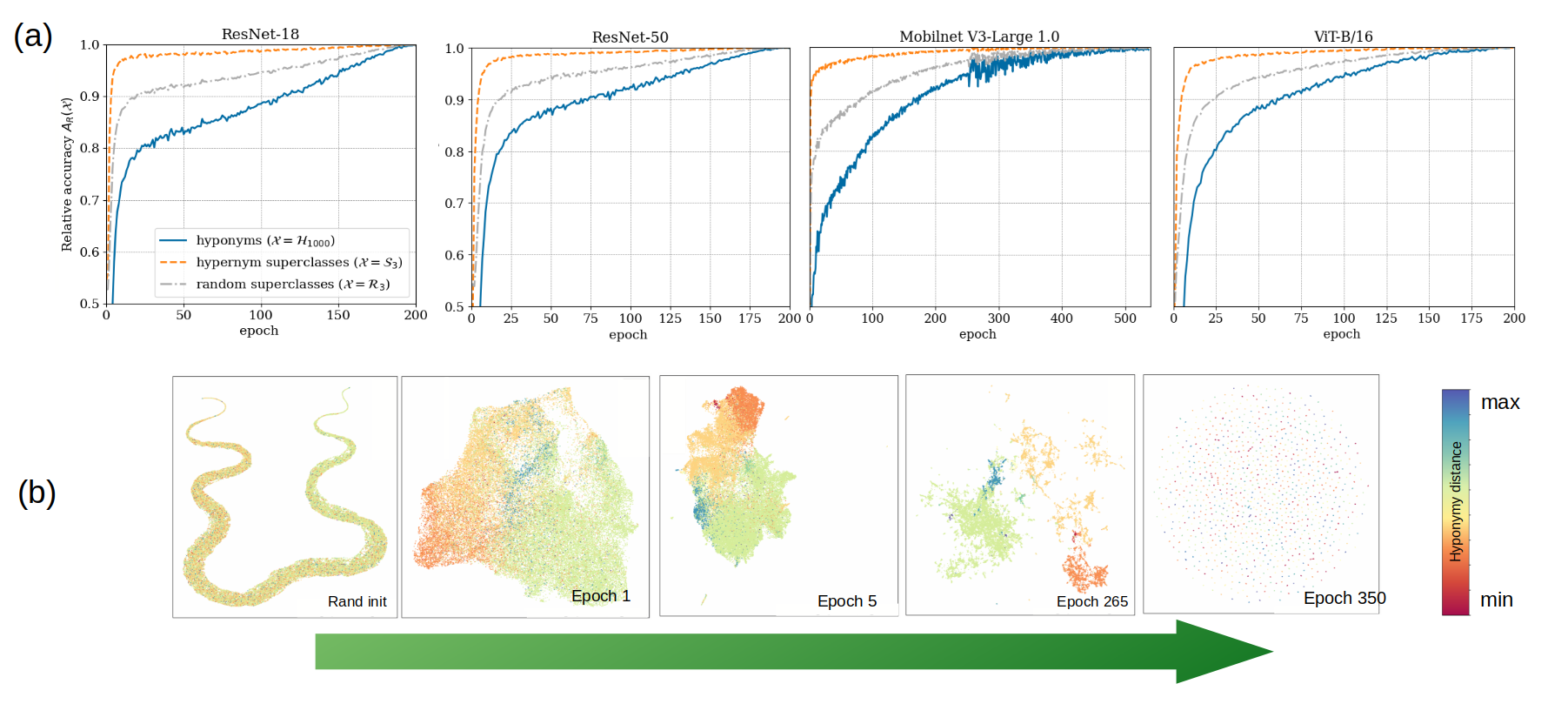}
  \caption{\label{fig_demo}Hypernym bias. (a) Relative accuracy during training
    for ResNet-18, ResNet-50, ViT-B/16, and MobileNet-V3. Hypernyms reach near
    maximum recognition accuracy on early epochs. (b) UMAP embeddings of
    ResNet-152 penultimate layer features in neural collapse settings. Color
    shows hyponymy distance relative to anchor class (<tabby cat>). The dynamics
    can be interpreted as top-to-bottom hierarchical label clustering}
\end{figure*}

In this work we study dynamics of the classifier learning process through the
lens of evolving hierarchy of object categories.
According to our experiments we argue that in the classification task the
network tend to learn relations between high-level categories (hypernyms) on
early stages, while more specific (hyponyms) are learned in the later training
epochs; terminal phase is characterized by removing hypernymy relations from the
features of the layer. By analogue we call this tendency a hypernym bias, as it
can be nicely interpreted as manifestation of simplicity bias: classes of same
hypernym share frequent (simple) features, therefore neurons responsible for
their detection are activated more often and trained faster.
The neural collapse state can be interpreted as final state of the label
clustering process in the specific layer: every cluster contains only a single
label, and information about labels' relations is removed from features. The
basic intuition about the phenomenon we study in this paper can be grasped from
Figure \ref{fig_demo}.\footnote{Video of UMAP embeddings evolution is available
  \href{https://doi.org/10.5281/zenodo.14851988}{here}.}

From \figurename~\ref{fig_demo}b training can be interpreted as top-to-bottom
hierarchical label clustering that starts by increasing distance between large
hypernyms\footnote{For example, on the 1st epoch features associated with close
  to anchor labels (red, orange and yellow dots) are close but not separated from
  each other} and neural collapse at the end of training, which indicates the
formation of singleton clusters containing only a single label, and thus
removing hyponymy relations from the feature space.

In this work:
\begin{itemize}
\item We provide empirical evidence that on early epochs of training classifiers
  tend to prioritize learning decision boundaries that distinguish superclasses
  of labels that unite large hypernyms, and thus learning can be seen from the
  perspective of hierarchical clustering of labels.
\item We assess the evolution of class hierarchical structure of the manifold. By
  comparing within manifold and hyponymy distances we show that clusters of
  labels that network learn are aligned with semantic dataset (WordNet) until
  the early stages of training.
\item We show that in hypernym label spaces some aspects of neural collapse
  manifest earlier than in original (hyponyms) label spaces.
\end{itemize}


\section{RELATED WORKS}

\textbf{Hierarchical classification}. Errors in trained classifiers tend to occur
more frequently between closely related classes, such as different species of
animals or various types of vehicles. This pattern is evident in the
block-diagonal structure of the confusion matrix, provided that the classes are
properly sorted. Initially, the ImageNet dataset was organized according to the
lexical database of semantic relations WordNet \citep{fellbaum2010wordnet} under
assumption that it will help to build new efficient classification algorithms
\citep{russakovsky2015imagenet}. Since then lots of works tried to incorporate
hierarchy bias into the training process of deep neural networks: either in
architecture \citep{hinton2015distilling, Malashin2016, xiao2014error} or in
loss function \citep{redmon2017yolo9000, brust2019integrating}. Some authors
reported promising results. However these findings are often derived in
nonoptimal training settings \citep{brust2019integrating} or from closed
datasets, which complicates fair comparison (e.g. \citep{hinton2015distilling}
as example). It is also well-known that, in few-shot leaning settings removing
hierarchically related classes from the pretraining process significantly
impacts the performance on these classes \citep{vinyals2016matching}. This
suggests that features of the network inherently capture hierarchical
relationships. \textit{We believe our result provide insight into why
  hierarchical structure is not used in state-of-the-art solutions}.

\textbf{Frequency Principle}. The Frequency Principle (FP) or Spectral bias is a
phenomenon in neural network training, independently discovered in
\citep{rahaman2019spectral} and \citep{xu2019frequency}, which states that the
low-frequency components of the target function are learned first by the neural
network. The principle is formulated for the multidimensional spectrum of data
(where the number of dimensions equals the number of elements in the input
feature vector). However, its interpretation is often associated with
one-dimensional (time sequences) or two-dimensional (images) spaces
\citep{xu2024overview}.

Theoretical justification of FP remains challenging \citep{xu2024overview} and
its applicability may be limited in some cases \citep{zhang2020rethink, wang2023neural}.
Nevertheless, FP serves as a valuable tool to think about training dynamics,
providing intuitive interpretations for existing results. For example, deep
image prior (DIP) \citep{ulyanov2018deep} leverages randomly initialized network
to restore image. It can be explained more generally by FP: by inherently
learning low-frequency components before high-frequency details, neural networks
used in DIP naturally reconstruct the essential structures of images first.
\cite{beyazit2024inductive} use FP to explain the poorer performance of neural
networks on tabular data compared to other methods (such as decision trees)
\citep{shwartz2022tabular}. \cite{benjdiraa2023guided} suggest to use high-pass
filter before calculating the similarity metric between the reconstructed and
reference images to improve reconstruction.



\textit{Interpreting hypernym bias as a manifestation of Frequency Principle is
  complicated} by two factors: a) the need of Fourier Transform computation of
high-dimensional data, which suffers from the curse of dimensionality and b)
challenges in applying the principle to naturally unordered datasets (which is
true for image classification).


\textbf{Simplicity bias}. \cite{arpit2017closer} have shown that simple patterns
are learned first before the network transitions into the noise memorization
phase. Simple patterns are associated with those that frequently occur across
many examples. \cite{arpit2017closer} compare learning dynamics on real data
with dynamics on noise (where no patterns are present). They suggest that
datasets contain a large number of simple examples with common patterns, which
are learned during the first epoch, after which the network spends most of its
time learning more complex examples. \cite{hu2020surprising} theoretically and
practically demonstrate that during the initial phase of training, a neural
network learns a linear function. In \citep{mangalam2019deep}, a more general
observation is experimentally demonstrated: during the early iterations,
networks learn to correctly recognize examples that can be distinguished by
shallow classifiers (such as SVMs or Random Forests). \cite{zhang2021linear}
emphasize that this property of neural networks helps prevent overfitting, even
in conditions of significant overparameterization. Therefore, early stopping can
be considered a method to prevent the learning of high-frequency noise
components in individual signal examples. Simplicity bias sometimes is
considered to be a pitfall \citep{shah2020pitfalls,pezeshki2022dynamics},
because it allows learning spurious correlations (shortcuts) and harms
generalization. These features are often not useful for generalization and are
referred as shortcuts \citep{geirhos2020shortcut, wang2022frequency}. A separate
research direction focuses on finding ways to reduce false correlations and
studying learning dynamics in their presence \citep{pezeshki2022dynamics,
  qiu2024complexity}.

The Frequency Principle and Simplicity Bias are closely related; the latter can
be thought of as more general but less quantifiable of the former. \cite{xu2021deep}
demonstrate that, in regression tasks, low frequencies carry the majority of the
information needed for signal reconstruction and are easier to learn.


\textit{In simplicity bias terminology frequent features are called simple, and
  simplicity bias should manifest in accelerated learning of hypernym features
  common among many classes. This is what we robustly observe in neural networks
  trained in practical settings}.

\textbf{Neural collapse}. Neural collapse (NC) phenomenon was first described
by \cite{papyan2020prevalence} and then was further extended to broader settings
in \citep{dang2024neural}, \citep{fang2021exploring} and other works. Neural
collapse is a state of the last layer of the neural classifier, which can be
observed in the terminal phase of training. It is characterized by zero
within-class variability of features, when class means form vertices of the
Equiangular Tight Frame (ETF), and classification decision is similar to
prototypes \citep{snell2017prototypical}. While the ETF configuration is optimal
from the perspective of robustness to adversarial attacks
\citep{papyan2020prevalence}, its contribution to generalization is arguable, as
NC is observed only when accuracy on a balanced train set reaches 100\%
\citep{hui2022limitations}. Nevertheless knowledge about dynamics of the last
layer is helpful, for example, for designing few-shot learning algorithms, where
often only last layer is tuned \citep{yang2023neural}.

\textit{Neural collapse of penultimate layer can be seen as deeper
  representation of perfectly diagonal confusion matrix, which doesn't posses
  any information about hyponymy relations. From the perspective of hypernym
  bias NC is the final stage of top-to-bottom hierarchical label clustering
  process}.

\section{METHODOLOGY}
\label{sec_methodology}

We base our conclusions from experiments with ImageNet dataset, which classes
are organized according to lexical database of semantic relations WordNet
\citep{fellbaum2010wordnet}. Validation of these conclusions on additional
datasets will be presented later in the paper. WordNet defines a graph where the
nodes are synsets (sets of synonyms), and the edges represent the
hypernymy-hyponymy relationships. Examples of synsets are: \{car, automobile\}
and \{small, compact\}. Synset \{dog\} is a hypernym relative to the synsets
\{bulldog\} and \{spitz\}, which are its hyponyms. It is important to note that
the hyponymy relationship is not the only one; for instance, meronymy
(part-whole, e.g., \{wheel\}-\{car\}), troponymy (one expressing an aspect of
the other, e.g., \{whisper\}-\{speak\}), and others relationships are often
considered. While it's arguably impossible to determine the relationships fully
and consistently, if we view hypernym bias through the lens of simplicity bias
\citep{arpit2017closer}, the imperfect nature of these relationships should not
negate the phenomenon. This is because hypernyms tend to share features across
multiple hyponyms.

For first series of experiments we used simplified to a tree WordNet graph
\citep{russakovsky2015imagenet}.
This allows as to interpret hyponym classifier as unambiguous greedy classifier
of hypernyms. For manifold analysis, we utilize complete WordNet graph, which
permits multiple parent relationships. To avoid confusion we refer to the
simplified version as WordNet tree, while the full version is referred to as
WordNet graph.

\subsection{Greedy hypernym classification}
\label{section:greedy_classifier}

\textbf{Label spaces}. Let $\mathcal{H}_{N_H} = \{h\}^{N_{\mathcal{H}}}$ be the
set of $N_{\mathcal{H}}$ hyponyms (classes), and
$\mathcal{S}_{N_s}=\{s\}^{N_{\mathcal{S}}}$ be the set of $N_{\mathcal{S}}$
hypernyms (superclasses), where each superclass $s_i=\{h\}^{N_{s_i}}$ is defined
such that
\begin{equation}
\label{eq_superclass_space_property}
\bigcup_{s \in \mathcal S} s = \mathcal{H}_{N_h} \quad \text{and} \quad s_i \bigcap s_j = \emptyset
\quad \forall s_i, s_j \in \mathcal{S}, \quad i \neq j.
\end{equation}
In the case of ImageNet, we have $N_H= 1000$. For simplicity we omit subscript
notations indicating the sizes of sets where possible. A standard classifier
network $f(x;\theta)$ equipped with softmax, maps image $x$ to a probability
distribution over hyponyms (classes) $h \in \mathcal{H}$ based on the current
weights $\theta$. The training objective utilizes the cross-entropy loss
function:
\begin{equation}
\mathcal{L}_{CE} = - \sum_{h \in \mathcal{H}} \log f_h(x;\theta) y_h(x),
\end{equation}
where $y_h(x)$ represents true probability that $x$ belongs to $h$ (0 or 1 in
one-hot encoding), and $f_h(x;\theta)$ is the estimated probability of the same.

We consider $f(x;\theta)$ as a function that also maps an image to different
label spaces greedily (not considering confidence level). We denote labels and
label predictions with $\hat{}$ symbol. The predicted hyponym is then given by:
\begin{equation}
  \hat{f}_{\mathcal{H}}(x;\theta) = \text{argmax}_{h \in \mathcal{H}} f_h(x;\theta),
\end{equation}
and the true label $\hat{y}_{\mathcal{H}}(x)=\text{argmax}_{h \in \mathcal{H}}
y_h(x)$

We explore manifestation of hypernym bias through the convergence dynamics by
introducing a trivial mapping $\mathcal{T}_{\mathcal{H} \to \mathcal{S}}:
\mathcal{H} \to \mathcal{S}$, which maps each hyponym to its parent hypernym at
a specified level in the WordNet tree. The predicted hypernym then:
\begin{equation}
  \hat{f}_{\mathcal{S}}(x;\theta) = \mathcal{T}_{\mathcal{H} \to \mathcal{S}}(\hat{f}_{\mathcal{H}}(x;\theta)),
\end{equation}
\begin{equation}
\hat{y}_{\mathcal{S}}(x) = \mathcal{T}_{\mathcal{H} \to \mathcal{S}}(\hat{y}_{\mathcal{H}}(x)).
\end{equation}
In our experiments $\mathcal{T}_{\mathcal{H} \to \mathcal{S}}$ is implemented as
simple reverse traversal of the WordNet tree starting from
$\hat{f}_{\mathcal{H}}$; it returns the first match with any element of
$\mathcal{S}$.


To evaluate the specific impact of hypernym relationships in the WordNet tree, we
introduce $\mathcal{R}$, a randomly generated label space that is isomorphic to
$\mathcal{S}$. Isomorphism here means that there is a one-to-one correspondence
between labels in $\mathcal{R}$ and $\mathcal{S}$, and the sizes of the
corresponding superclasses are identical. Formally there exists bijection
$g:\mathcal{R} \to \mathcal{S}$, such that $\forall r \in \mathcal{R} \quad
\exists s \in \mathcal{S}: |r| =|g(r)|$. This ensures that any differences in
network performance on $\mathcal{R}$ and $\mathcal{S}$ can be attributed to the
semantic structure encoded in $\mathcal{S}$. For example, while $\mathcal{S}$
might group classes into "Animals", "Artifacts", and "Others", $\mathcal{R}$
would have three groups of the same sizes but with randomly assigned classes.



Given that $\mathcal{T}_{\mathcal{H} \to \mathcal{R}}$ and
$\mathcal{T}_{\mathcal{H} \to \mathcal{S}}$ are predefined and not learned, we
can assume that the same network $f(x;\theta)$ maps each image to three label
spaces simultaneously without interfering training process:
\begin{equation}
\hat{f}_{\mathcal{H}}(x;\theta) \to \mathcal{H},
\hat{f}_{\mathcal{S}}(x;\theta) \to \mathcal{S},
\hat{f}_{\mathcal{R}}(x;\theta) \to \mathcal{R}.
\end{equation}
Corresponding ground truths labels are $\hat{y}_{\mathcal{H}}(x),
\hat{y}_{\mathcal{S}}(x), \hat{y}_{\mathcal{R}}(x)$. Thus we can study accuracy
metrics in each label space.

\textbf{Metrics}. Let the test dataset $D=\{x_i\}_{i=1}^{N_D}$ contain $N_D$
images for which label function $\hat{y}$ is defined. The accuracy of
classification in label space $\mathcal{X}$ after $t$ epochs is estimated as:
\begin{equation}
A(\mathcal{X}, t) = \frac{1}{N_D} \sum_{i=1}^{N_D} \mathbb{I}(\hat{f}_{\mathcal{X}}(x_i; \theta_t) = \hat{y}_{\mathcal{X}}(x_i)).
\end{equation}
We normalize $A(\mathcal{X}, t) \in [0,100]$.

Given the accuracy of recognizing hyponyms $A(\mathcal{H},t)$, accuracy of
recognizing random superclasses $A(\mathcal{R},t)$ can be estimated
theoretically. We derive and validate exact formula in
\appendixname~\ref{appendix_methodology}.

Absolute accuracy can be misleading indicator of convergence since it varies a
lot (as recognizing superclasses is inherently easier due to higher probability
of guessing correctly). To address this, we introduce relative accuracy
$A_R(\mathcal{X},t)$ at epoch $t$ for assessment. Let $T=\text{argmax}_{t}
A(\mathcal{X},t)$, then relative accuracy is defined as:
\begin{equation}
\label{eq_relative_acc}
A_R(\mathcal{X},t) = \frac{A(\mathcal{X},t)}{A(\mathcal{X},T)},
\end{equation}
which ensures that $A_R(\mathcal{X},t) \in [0,1]$.

Expression (\ref{eq_relative_acc}) does not account for different probabilities
of random guessing that depends on superclass sizes distribution. To address
this, we also introduce the relative gain in accuracy, $G_R(\mathcal{X},t)$,
defined as:
\begin{equation}
\label{eq_relative_gain}
G_R(\mathcal{X},t) = \frac{A(\mathcal{X},t) - B(\mathcal{X})}{A(\mathcal{X},T) - B(\mathcal{X})},
\end{equation}
where $B(\mathcal{X})$ is the accuracy of a random guess, taking into account
sets imbalance. This baseline accuracy can be estimated as $B(\mathcal{X}) =
\mathbb{E}A(\mathcal{R},0)$, i.e. expected accuracy before
training. 
In the next section we show that estimates of accuracy gain for random
superclasses $G_{R}(\mathcal{R})$ and for hyponyms $G_{R}(\mathcal{H})$ are
identical, allowing us to measure impact of hypernymy relation fairly by
comparing $A_{R}(\mathcal{R})$ against $A_{R}(\mathcal{S})$.


Though observation of accelerated formation of hypernym features from the
perspective of different accuracy metrics are very practical and intuitive
manifestation of hypernym bias, these metrics are not integral (need arbitrary
choices of hypernym level), and importantly they doesn't generalize to describe
features of deeper layers.

\subsection{Assessing evolution of class hierarchical structure of the manifold}
\label{section:manifold}
Assuming that features lie near low-dimensional manifold, we evaluate how well
this feature manifold aligns with WordNet graph on each epoch.
Our approach has three major steps: a) define the distances between classes in
feature space, b) define distances between classes in WordNet graph and c)
compare two distance matrices via cophenetic correlation coefficient.

\textbf{Graph in feature space}. To estimate distances between feature sets
within manifold we adapt approach from \citep{jin2020quantifying}. First we
sample $2 \times K$ training examples per each of $C$ classes, and divide them
into non-overlapping query and support sets $Q$ and $S$:
\begin{equation}
Q = \left\{ U^c = \left\{ u^c \right\}^{K} \right\}^{C}, \; S = \left\{ V^c = \left\{ v^c \right\}^{K} \right\}^{C}, \\
\end{equation}
where $U^c \cap V^c = \emptyset, \; \forall c \in [0, C]$.
Assuming suitable metric, we define distance $d_f(u^{c_i}, V^{c_j})$ between
query feature of class $c_i$ and support set $V^{c_j}$ of another class $c_j$ as
minimum individual distance:
\begin{equation}
  d_f({c_i}, V^{c_j}) = \min_{v \in V^{c_j}} d_f(u^c, v).
\end{equation}

Next we measure probability that distance between query point of class $c_i$ and
support set $V^{c_j}$ is less than predefined radius $r$:
\begin{align}
  P_r(c_i, c_j) &= P \left( d_f(u^{c_i}, V^{c_j}) < r \right) = \\
  &= \mathbb{E}_{u \sim U^{c_i}} \left[ \mathbb{I}\{ d_f(u, V^{c_j}) < r \} \right],
\end{align}
where $\mathbb{I}\{ d_f(u, V^{c_j}) < r \}$ is the indicator function.

Finally we estimate similarity of two classes as:
\begin{equation}
\rho(c_i, c_j) = \frac{1}{r_{\text{max}}} \int_0^{r_{\text{max}}} P_r(c_i, c_j) \, dr,
\end{equation}
where $r_{max}$ is maximum radius.

The value $\rho(c_i, c_j)\in [0,1]$ represents mutual cover when $i \neq j$ and
self-cover when $i=j$, as described by \cite{jin2020quantifying}. Similarity
$\rho(c_i, c_j) = 1$ indicates that every query feature $u^{c_i}$ is within zero
distance to \textit{at least one} feature $v^{c_j}$ in support set of the class
$j$, low values indicate that there are many query features that are far from
\textit{any} features in $V^{c_j}$. We construct similarity matrix $A = \left[
  a_{ij} \right], \; \text{where} \; a_{ij} = \rho(c_i, c_j)$ and treat its
elements as probability that edge between i-th and j-th node of the graph
exists.
Transformation from similarity to a distance matrix $D_f$ is straightforward:
\begin{equation}
D_F = \left[d_F(i,j)] \right]=1-A.
\label{eq:graphdistance}
\end{equation}

\textbf{Label distance in WordNet graph}. Distance matrix between class labels
according to WordNet structure can be defined as:
\begin{equation}
  D_W = \left[ d_W(i,j) \right],
\end{equation}
where $d_w(i, j)$ is the shortest path between synset $i$ and $j$ in the graph.
With this definition we can use full (not tree) WordNet graph.

\textbf{Graph comparison}. To compare graphs we utilize Cophenetic Correlation
Coefficient:

\begin{equation}
  \text{CCC} = \frac{\sum_{i < j} \tilde{d}_W(i,j) \tilde{d}_F(i,j)}
  {\sqrt{\sum_{i < j} \tilde{d}_W(i,j)^2 \sum_{i < j} \tilde{d}_F(i,j)^2}},
\end{equation}

\begin{equation}
  \tilde{d}_W(i,j) = d_W(i,j) - \overline{D_W}, \quad
  \tilde{d}_F(i,j) = d_F(i,j) - \overline{D_F},
\end{equation}

where   $\overline{D_F}$ and $\overline{D_W}$ are the mean of all pairwise
distances in the original distance matrices $D_F$ and $D_W$. $CCC \in [-1,1]$
measures linear correlation of two distance matrices.

\begin{figure}[h]
  \centering
  \includegraphics[width=8cm]{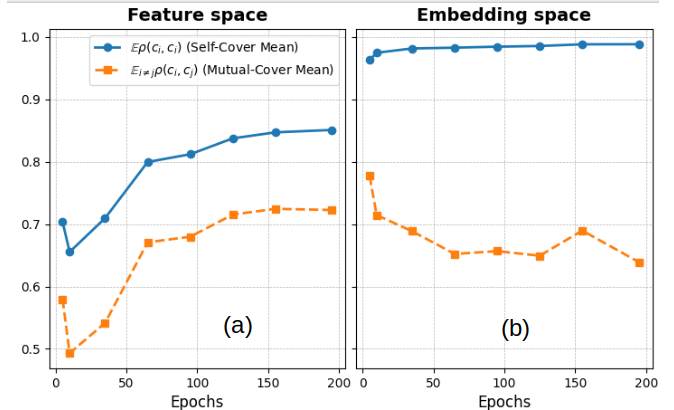}
  \caption{\label{fig_cover_dynamics} Mean of mutual- and self-covers of the
    features of penultimate ResNet-50 layer in the course of training: original
    feature space (a) and UMAP embeddings space (b)}
\end{figure}

\textbf{Metric}.
Examples are not distributed uniformly in feature space in the course of
training: fixed volume contains different number of sample points and distance
should be adjusted accordingly. As can be seen from
\figurename~\ref{fig_cover_dynamics}a with the use of Euclidean metric in
feature space there is no trend to increase mutual to self-cover ratio. This
suggests the potential issues in the measurement of similarity, as the Euclidean
metric likely captures the properties of the entire manifold rather than
specific characteristics of individual class manifolds and their mutual
relationships.
In this work we address the issue by utilizing the formally grounded UMAP
algorithm \citep{mcinnes2018umap}, which approximates a manifold where the data
is uniformly distributed and effectively normalizes distances according to local
connectivity before embedding optimization. We transform the feature space into
an M-dimensional embedding space, where the usage of Euclidean metric $d(u,v)
=\|u-v\|_2$ is justified. The dynamics of self-to-mutual cover ratio in UMAP
embedding space depicted in \figurename~\ref{fig_cover_dynamics}b, exhibit
expected behavior of an increasing cover difference.

\subsection{Neural collapse in hypernym label space}

Neural collapse \citep{papyan2020prevalence} is a phenomenon that is observed in
penultimate layer of the classifier when recognition accuracy on the training
set reaches 100\%. Neural Collapse is characterized by four manifestations in
the last-layers weights and last-layer activations: within-class variability
collapse (NC1), convergence to simplex ETF (NC2), convergence to self-duality
(NC3) and simplification to nearest-class center (NC4). We give brief overview
of neural collapse in \appendixname~\ref{appendix_methodology}. Originally NC is
defined on the label space $\mathcal{H}$, where training loss is estimated.
Assuming hypernym bias we expect at least some properties of neural collapse in
label spaces of hypernyms emerge earlier. We replicate calculations as in
\citep{papyan2020prevalence} but in different label spaces after applying
$\mathcal{T}_{\mathcal{H} \to \mathcal{R}}$ or $\mathcal{T}_{\mathcal{H} \to
  \mathcal{S}}$, described in Section \ref{section:greedy_classifier}. To
compute NC2-NC4 we must consider that the weight matrix
$\mathbf{W}=[\mathbf{w}_c]$ contains C weights rows, consistent with the number
of labels. We interpret each row $\mathbf{w}_c$ as prototype
\citep{snell2017prototypical} of its own hyponym. Assuming linear relationships
to obtain prototypes of hypernyms $s \in \mathcal{S}$ we can average columns
under hypernym relation. Thus rows of the resulting weight matrix can be
calculated as $\mathbf{w}_{s} \triangleq \text{Ave}_{c} \{\mathbf{w}_{c}\}, \; c
\in s$.

\section{EMPIRICAL RESULTS}
\label{sec_empirical_res}
\subsection{Experimental setup}
\textbf{Top level hypernyms}. Top level of the WordNet tree contains 9 synsets.
we expect manifestation of hypernym bias should be more pronounced in large
hyperyms. Since top-level synsets are unevenly saturated with classes, we
grouped them into synsets sets and used as top level hypernym superclasses; for
clarity, most of the results are reported with respect these label spaces: 1)
hyponyms $\mathcal{H}$ : 1000 classes used for training, 2) hypernym
superclasses $\mathcal{S}_3$: three superclasses formed from 522 (artifacts),
398 (animals), and 80 (others) classes according to the WordNet tree top level
synsets 3) random superclasses $\mathcal{R}_3$: three superclasses formed from
80, 522, and 398 randomly selected classes.

\textbf{Lower level hypernyms}. We demonstrate the robustness of the observed
phenomenon to different specifications of $\mathcal{S}$ with the use of
different levels of hypernymy tree.

\textbf{Architectures and hyper parameters}. In experiments we used the
ResNet-50, ResNet-18 \citep{he2016deep}, ViT-B/16 \citep{Dosovitskiy2020} and
MobileNet-V3 \citep{howard2019searching} architectures. All models were trained
with the use of augmentation, starting from randomly initialized weights. Except
for Vit-B/16, the training parameters gave results comparable with the state of
the art results for these architectures. For neural collapse analysis, we
carefully reproduced all the parameters used by \cite{papyan2020prevalence} and
trained ResNet-152 for 350 epochs. We did not perform a learning rate search and
fixed it to be 0.1. For the greedy hypernym classifier we report metrics on
validation set, while neural collapse is evaluated on the train set. To compute
$\text{CCC}$ we embedded features into a 10-dimensional feature space.

Details on superclass formation with the use of WordNet tree and training
parameters can be found in Appendix \appendixname~\ref{appendix_exp_details}.

\subsection{Results}

\begin{figure}[h]
\centering
\includegraphics[width=7cm]{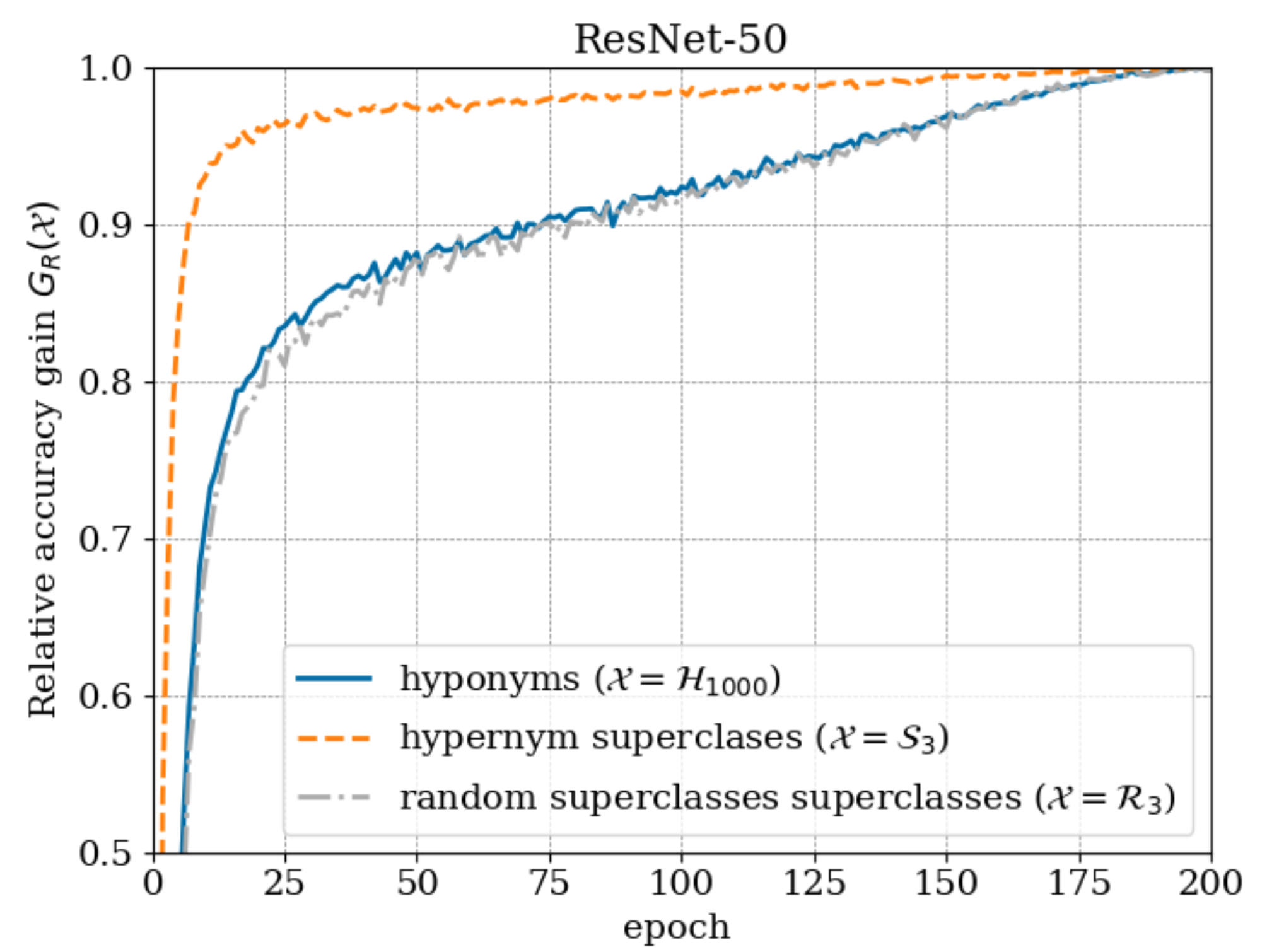}
\caption{\label{figure2} Relative accuracy gain during training of ResNet-50.}
\end{figure}

\textbf{Greedy hypernym classifier}. \tablename~{\ref{tab_rec_acc}} shows final
absolute accuracy, and \figurename~\ref{fig_demo}a displays relative accuracy
curves in hyponym, hypernym and random superclass label spaces. Across all
architectures during initial epochs of training relative accuracy for hypernyms
increases significantly faster than for the rest of considered label spaces.
\begin{table*}[htbp]
  \caption{\label{tab_rec_acc} Top-1 recognition accuracy for different label
    spaces}
\centering
\begin{tabular}{lrrrr}
  Set & ResNet-50 & ResNet-18 & ViT-B/16 & MobileNet-V3\\
  \hline
  hyponym,  $A(\mathcal{H}_{1000})$ & 79 & 69.8 & 75.6 & 75.6\\
  hypernym, $A(\mathcal{S}_3$)& 97.7 & 96.3 & 97.1 & 97.2\\
  random, $A(\mathcal{R}_3)$  & 87.6 & 83 & 86.3 & 86.3\\
\end{tabular}
\end{table*}
From the table, it is evident that the recognition accuracy of superclasses
formed with the use of the WordNet tree is significantly higher compared to
random superclasses of the same size. However, the final hypernym classification
accuracy strongly depends on the size of the network. For instance, ResNet-50
outperforms ResNet-18 by 61\% in terms of error reduction for hypernyms,
compared to a 43\% improvement for hyponyms. This should be taken into account
when using interpretation through the lens of simplicity bias. Also hypernym
accuracy never fully plateaus until the end of training. Figure~\ref{figure2}
shows relative accuracy gain computed according to (\ref{eq_relative_gain}) for
ResNet-50.
The metric behaves identically for random superclasses and hyponyms showing its
invariance to label imbalance. It clearly demonstrates accelerated dynamics of
learning hypernyms, but more importantly it justifies drawing conclusions about
hypernym vs hyponym training dynamics by comparing $A_R(\mathcal{S}_n)$ against
$A_R(\mathcal{R}_n)$ (instead of $A_R(\mathcal{H}_{1000})$).

\begin{figure}[h]
\centering
\includegraphics[width=8cm]{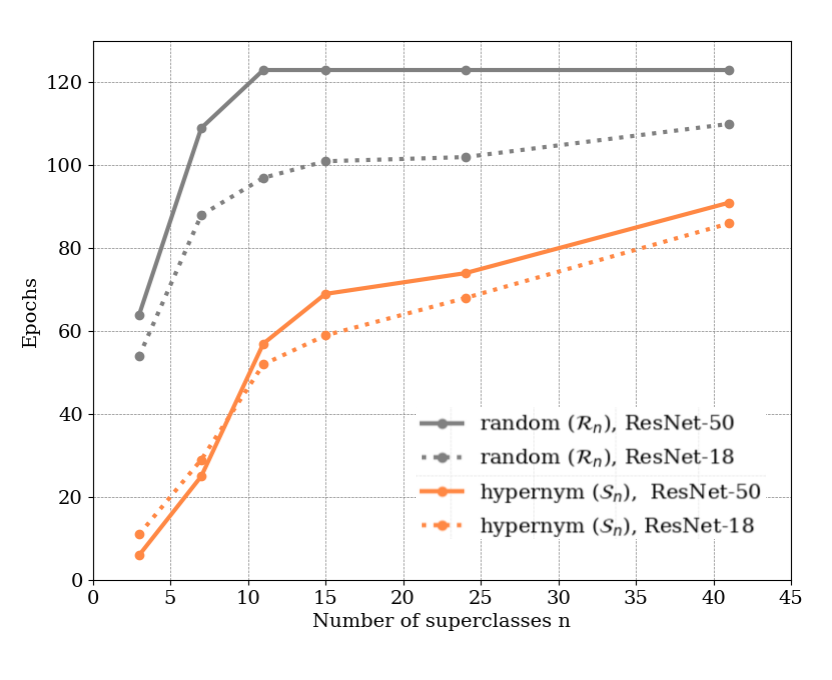}
\caption{\label{fig_many_superclasses} 95\% accuracy convergence period (in
  epochs) for different number of hypernyms}
\end{figure}

As an indicator of the training convergence with respect to number of hypernyms
we considered the number of epochs required to reach 95\% of the maximum
accuracy. The results are depicted in Figure \ref{fig_many_superclasses}.

As observed, the recognition accuracy of hypernyms converges significantly
faster, although smaller hypernyms tend to take longer to reach convergence. For
instance, ResNet-50 training reaches 95\% accuracy in 6 epochs for
$\mathcal{S}_3$, while for randomly formed superclasses $\mathcal{R}_3$, it
takes 64 epochs. Interestingly, the effect of accelerated hypernym learning is
more salient in larger ResNet-50 than in smaller ResNet-18. In contrast, the
convergence rate for hypernyms is approximately equal for both networks (orange
lines in Figure \ref{fig_many_superclasses}), whereas the features required for
hyponym classification consistently take longer to converge (gray lines on the
same graph) in ResNet-50.

Although our choice of grouping synsets within one WordNet level was arbitrary,
the absence of such grouping and the choice of a different convergence threshold
do not affect the conclusions obtained in this section.
More experiments, including experiments with animals hypernyms, different synset
grouping strategies can be found in
\appendixname~\ref{appendix_add_experiments}.

\begin{figure*}[h]
  \centering
  \includegraphics[width=16cm]{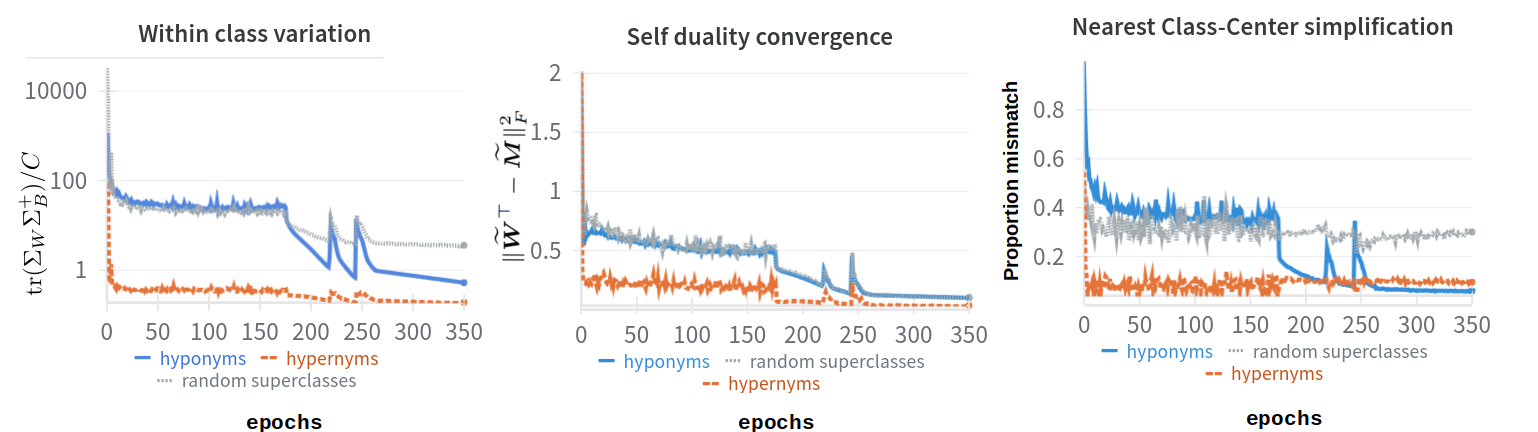}
  \caption{\label{fig_nc_graphs} Estimation of NC1, NC3 and NC4 properties of neural
  collapse in different label spaces}
\end{figure*}

\textbf{Neural collapse}. In \figurename~\ref{fig_nc_graphs} we plot graphs
demonstrating that some properties of neural collapse are observed in
$\mathcal{S}_3$ hypernym label space on early iterations of training.
UMAP embeddings of penultimate features are shown in Figure \ref{fig_demo}. We
provide all the graphs as well as more UMAP visualizations in the
\appendixname~\ref{appendix_add_experiments}.

\subsection{Manifold hierarchical structure, layer-wise and data frequency
  analysis}

\begin{figure}[h]
  \centering
  \includegraphics[width=8cm]{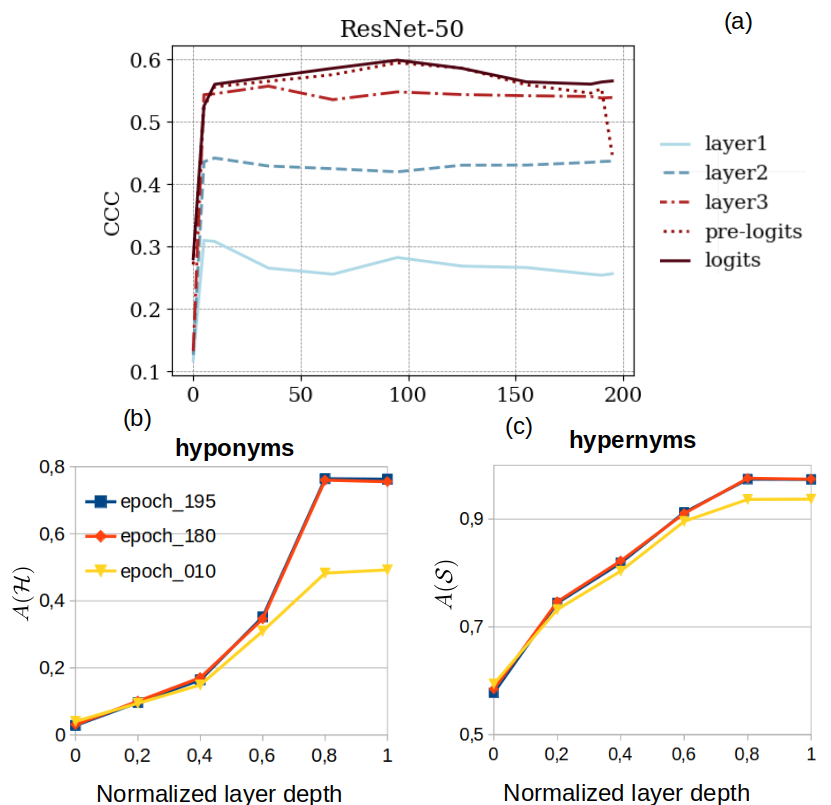}
  \caption{\label{fig_final_figure} Cophenetic correlation coefficient for
    WordNet graph and mutual covers distance matrix (a). Linear probing of
    several checkpoints of ResNet-50 in hyponym (b) and hypernym (c) spaces }
\end{figure}

\textbf{Alignment with WordNet graph and layer-wise dynamics}. It is well-known
that layers of convolutional neural networks converge faster than the entire
neural network \citep{raghu2017svcca, wang2023egeria}. Residual skip connections
can possibly help transmit less distorted low-level feature information to the
upper layer responsible for decision-making. Thus, layer-wise dynamics can
possibly be related to hypernym bias: hypernyms can be recognized without use of
high level features. We apply framework developed in Section
\ref{section:manifold} to evaluate how well feature manifold of different levels
aligns with WordNet graph on each epoch. The results presented in
\figurename~\ref{fig_final_figure}a deny suggested hypothesis.

As can be seen from the figure WordNet graph and manifold distances linear
alignment rapidly grows in first 5 epochs. Maximum correlation of 0.6 is
achieved in the middle of the training by top layers and starts to decay after
that. Bottom layers consistently show lower values. These conclusion is in
coherence with the results of using linear probes \citep{alain2016understanding}
in hypernym label spaces (Figure \ref{fig_final_figure}c), which show layers'
redness for hypernym classification. We observe that probes from higher levels
consistently and similarly provide better accuracy for both hyponym and
hypernym spaces. In no experiment do we observe that hypernym classification based on
intermediate layers can be performed with accuracy comparable to classification
based on the higher layer. Thus, we cannot conclude that knowledge sufficient
for recognizing hypernyms is contained in the first layers.

\textbf{Frequency analysis}. The Frequency Principle \citep{xu2019frequency} is
often linked to image frequencies, suggesting that hypernyms could be recognized
using low frequencies, which are learned early in training, while hyponyms might
require high frequencies learned later. We tested this hypothesis and conclude
that metrics in hypernyms label spaces do not exhibit a lesser dependency on
high frequencies in the first phase of training.

These and some other experimental results, the description of the probing
methods are given in \appendixname~\ref{appendix_add_experiments}.

\subsection{Generalization}
There are no formal grounds to believe that the results should generalize beyond
the conducted experiments. To address this concern, we performed additional
experiments to further validate our findings.

\textbf{CIFAR-100}. CIFAR-100 \citep{krizhevsky2009learning} comprises 60~000
$32 \times 32$ images distributed across 100 classes, which are further grouped
into 20 superclasses. Let $\mathcal{S}_{20}$ represent the hypernym label space
(superclasses) and $\mathcal{R}_{20}$ denote a random label space isomorphic to
$\mathcal{S}_{20}$.

We trained ResNet-18 and achieved 77\% validation accuracy in hyponym label
space.

We employed the greedy classifier model described in Section 3.1 and observed
accelerated growth of $A_R(\mathcal{S}_{20})$ compared to
$A_R(\mathcal{R}_{20})$, as expected.

\textbf{DBPedia}. DBPedia \citep{lehmann2015dbpedia} is a text classification dataset containing 630,000 samples. It
includes 14 classes, which we grouped into 5 hypernyms ($\mathcal{S}_{5}$) based
on the first level of the DBPedia ontology. We fine-tuned a pretrained BERT-base
model, and the test errors were as follows: 0.48\% for random superclass labels,
and 0.37\% for hypernym labels. The relative accuracy curves demonstrate that
the difference between $A(\mathcal{R}_5)$ and $A(\mathcal{S}_5)$ emerges during
the early iterations, as predicted. Hypernym bias is evident.

Further details on the CIFAR-100 and DBPedia experiments, including relative
accuracy curves, are provided in \appendixname~\ref{appendix_other_sets}.

\section{DISCUSSION AND FUTURE DIRECTIONS}

\textbf{Hierarchical classification}. To the best of our knowledge, the WordNet
graph has not been widely adopted in state-of-the-art classification algorithms.
Our experiments suggest that this may be because the hyponym-hypernym
relationships are implicitly learned by the network itself. Our results show
that when the loss function incorporates multiple cross-entropy losses from
different hierarchy levels (as in \citep{redmon2017yolo9000}), the
contributions of higher-level hierarchy components are significant only during
the early iterations of training, with a much lower impact in later stages.

For example, \cite{brust2019integrating} report similar accuracy metrics when
leveraging the hyponymy relation on CIFAR-100, but they observe the faster
convergence and a shift in the initial training phase. Our findings explain this
by showing that hypernym relations are not independently learned by the
network only in the first epochs of training. Moreover,
\cite{brust2019integrating} also report a notable increase in accuracy when
incorporating hierarchical relations during training on ImageNet. We attribute
this improvement to their use of the mean squared error loss function, which is
suboptimal for classification tasks.

\textbf{Future work}. Future research could extend the notion of hypernym bias to
other domains. For instance, the AudioSet \citep{gemmeke2017audio} has a
hierarchical organization of categories, such as "Human sounds" → "Speech" and
is well suited to the proposed framework.

Another promising area of research involves moving away from reliance on
predefined external class structures and instead learning class hierarchies
directly through data-driven hierarchical clustering. For example, the class
similarity matrix $A$ from Equation \eqref{eq:graphdistance} can be treated as
an adjacency matrix, reframing the problem into identifying hierarchical
communities in the graph, to which known algorithms can be applied. A
data-driven approach has two key benefits: (a) hierarchical clustering can
uncover more accurate and intuitive representations of class relationships,
potentially revealing latent patterns missed by external structures, and (b)
analyzing the training dynamics of classifiers in conjunction with these learned
hierarchies could provide valuable insights for designing more efficient and
adaptive clustering algorithms.


Investigating the theoretical connections between hypernym bias and other known
biases is a promising direction. Establishing these relationships would not only
deepen the theoretical understanding of hypernym bias but also bridge practical
findings with theoretical models, which often lack strong empirical validation.

\section{CONCLUSIONS}

The primary objective of our work was to establish that hypernym bias is a
consistent phenomenon across different neural network architectures trained in
practical settings, which other works on biases and hierarchical learning often
lack. We also aimed to introduce an experimental framework that provides a
systematic and quantifiable approach to studying biases in practical scenarios,
which can serve as a foundation for future research in this area. To the best of
our knowledge we are also the first to connect the training bias in the early
stage with neural collapse in the terminal training stage by estimating neural
collapse in different label spaces. Therefore the notion of hypernym bias can
may contribute to research of neural collapse. Our experiments, which did not
reveal parallels with existing phenomena, were designed to demonstrate that
hypernym bias cannot be trivially attributed to other known factors, such as
layer-wise training dynamics or data frequency biases. The lack of clear
parallels in these cases underscores that hypernym bias is a distinct phenomenon
requiring independent investigation.

\section{Aknowledgement}
The paper was prepared with the financial support of the Ministry of Science and
Higher Education of the Russian Federation, grant agreement No. FSRF-2023-0003,
``Fundamental principles of building of noise-immune systems for space and
satellite communications, relative navigation, technical vision and aerospace
monitoring''

\bibliographystyle{apalike}
\bibliography{hypernim_bias}

\section*{Checklist}
 \begin{enumerate}

 \item For all models and algorithms presented, check if you include:
 \begin{enumerate}
 \item A clear description of the mathematical setting, assumptions, algorithm,
   and/or model. [Yes, see Section \ref{sec_methodology}]
   \item An analysis of the properties and complexity (time, space, sample size)
     of any algorithm. [Yes, see Appendix~\ref{appendix_complexity}]
   \item (Optional) Anonymized source code, with specification of all dependencies, including external libraries.[No]
 \end{enumerate}

 \item For any theoretical claim, check if you include:
 \begin{enumerate}
   \item Statements of the full set of assumptions of all theoretical results. [Yes]
   \item Complete proofs of all theoretical results. [Yes, see Appendix~\ref{appendix_methodology}]
   \item Clear explanations of any assumptions. [Yes]
 \end{enumerate}

 \item For all figures and tables that present empirical results, check if you include:
 \begin{enumerate}
   \item The code, data, and instructions needed to reproduce the main experimental results (either in the supplemental material or as a URL). [No]
   \item All the training details (e.g., data splits, hyperparameters, how they
     were chosen). [Yes, see Section~\ref{sec_empirical_res}, Appendix~\ref{appendix_exp_details}]
   \item A clear definition of the specific measure or statistics and error bars
     (e.g., with respect to the random seed after running experiments multiple
     times). [Yes, see Section~\ref{sec_methodology}, Appendix~\ref{appendix_resid_error}]
   \item A description of the computing infrastructure used. (e.g., type of
     GPUs, internal cluster, or cloud provider). [Yes, see
     Appendix~\ref{appendix_complexity}]
 \end{enumerate}

 \item If you are using existing assets (e.g., code, data, models) or curating/releasing new assets, check if you include:
 \begin{enumerate}
   \item Citations of the creator If your work uses existing assets. [Yes]
   \item The license information of the assets, if applicable. [Not Applicable]
   \item New assets either in the supplemental material or as a URL, if applicable. [Not Applicable]
   \item Information about consent from data providers/curators. [Not Applicable]
   \item Discussion of sensible content if applicable, e.g., personally identifiable information or offensive content. [Not Applicable]
 \end{enumerate}

 \item If you used crowdsourcing or conducted research with human subjects, check if you include:
 \begin{enumerate}
   \item The full text of instructions given to participants and screenshots. [Not Applicable]
   \item Descriptions of potential participant risks, with links to Institutional Review Board (IRB) approvals if applicable. [Not Applicable]
   \item The estimated hourly wage paid to participants and the total amount spent on participant compensation. [Not Applicable]
 \end{enumerate}
\end{enumerate}

\appendix

\onecolumn


\section{METHODOLOGY}
\label{appendix_methodology}
\subsection{Relation of $A(\mathcal{R},t)$ and $A(\mathcal{H},t)$}
\label{appendix:theoretical_acc}

Let $f_{\mathcal{H}}(x; \theta_t)$ denote a classifier trained in the hyponym
label space \( \mathcal{H} \) at epoch \( t \). If \( A(\mathcal{H}, t) \)
represents the absolute classification accuracy in $\mathcal{H}$, then the
classification accuracy in a random superclass label space, $A(\mathcal{R}, t)$,
can be theoretically estimated based on the sizes of the subsets within
$\mathcal{R}=\{r\}$.

To see this, let $D=\{x_i\}_{i=1}^{N}$ be the dataset with $N$ examples and
$A(\mathcal{X}, t) \in [0,100]$ in every label space $\mathcal{X}$. Accuracy
$A(\mathcal{H}, t)$ is essentially an estimation of probability
$P_{\mathcal{H}}(x;\theta_t)$ that classifier $f$ assigns correct hyponym label
$\hat{y}_{\mathcal{H}}(x)$ to an image $x$:
\begin{equation}
  A({\mathcal{H},t}) / 100 \approx P_{\mathcal{H}}(x;\theta_t) = P (\hat{f}_{\mathcal{H}}(x;\theta_t) = \hat{y}_{\mathcal{H}}(x)),
\end{equation}
where $\hat{f}_{\mathcal{H}}(x;\theta_t)$ is the predicted label.

Assume that we use classifier $f$ to greedily predict the label in the random
superclass label space $\mathcal{R}$, accordingly to Section 3.1. Let
$\hat{f}_{\mathcal{R}}(x;\theta_t)$ be a superclass label that is predicted. The
probability $P(\hat{y}_{\mathcal{R}}(x)=r)$ that a randomly chosen image $x$
belongs to a superclass $r$ can be estimated according to the size of the
superclass in the dataset:
\begin{equation}
  P(\hat{y}_{\mathcal{R}}=r) \approx P_{r} \triangleq \frac{|D_r|}{N_D},
\end{equation}
where $D_r = \{x \in D \mid \hat{y}_{\mathcal{R}}(x)=r\}$ is a set of examples
with label $r$.

Since hyponyms are assigned to $r$ independently, prior probability $P_{R}$ that
greedy hypernym classifier predicts correct label of superclass $r$:
\begin{equation}
  P_{R}(\hat{f}_{\mathcal{R}}(x) = \hat{y}_{\mathcal{R}}(x)|\hat{y}_{\mathcal{R}}(x)=r) = P_r.
\end{equation}

The probability of assigning a correct label in the superclass label space given
a hyponym recognition probability $P_{\mathcal{H}}(x;\theta_t)$, can be
estimated by considering two possible events: (a) correctly classifying the
hyponym, or (b) misclassifying the hyponym but still assigning the correct
superclass label by chance.

\begin{equation}
\label{eq_prob}
P_{\mathcal{R}}(x;\theta_t) =
P_{\mathcal{H}}(x;\theta_t) + (1 - P_{\mathcal{H}}(x;\theta_t)) \sum_{r \in \mathcal{R}} P_r^2,
\end{equation}

\begin{figure}[h]
\centering
\includegraphics[width=8cm]{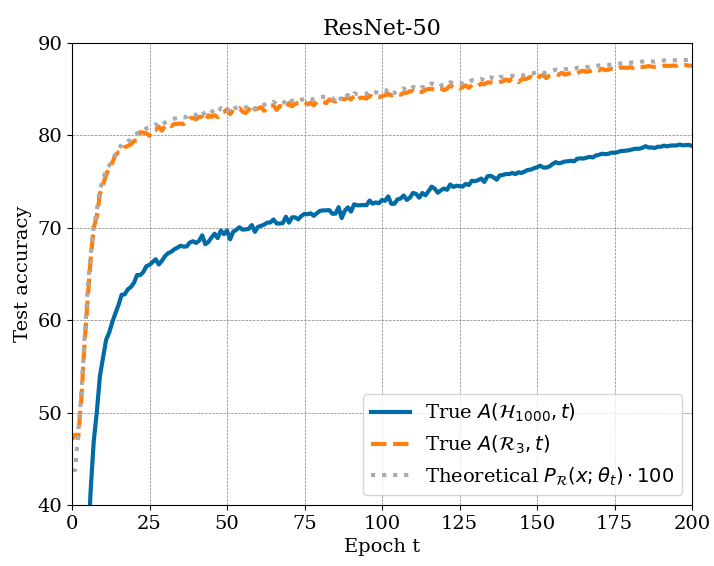}
\caption{\label{fig_formula_verification} Experimentally observed accuracy in
  random superclasses label space and theoretically estimation of the same,
  given accuracy in hyponym label space.}
\end{figure}

$P_{\mathcal{R}}(x;\theta_t)$ gives good estimate of $A(\mathcal{R}, t)$ in real
experiments (Figure \ref{fig_formula_verification}) except at the beginning of
training, due to the nonuniform behavior of the randomly initialized network.

\subsection{Neural Collapse}
\label{sec:nc_theory}
Here we briefly overview Neural Collapse (NC) proposed by
\cite{papyan2020prevalence}. Originally NC is defined only in hyponym label
space (used for training), so we extend it to the hypernym label spaces.

\textbf{Hyponym Label Spaces}. Let $\mathbf{h}_{i,c} \in \mathbb{R}^p$ be a
$p$-dimensional feature vector, where $c$ is one of $|\mathcal{X}|$ classes,
where $\mathcal{X}$ is the label space. In the original ImageNet hyponym label space
($\mathcal{X} = \mathcal{H}$) each class corresponds to a hyponym, so
$|\mathcal{H}|=1000$. The matrix $\mathbf{W} \in \mathbb{R}^{|\mathcal{H}|
  \times p}$ and vector $\mathbf{b} \in \mathbb{R}^{|\mathcal{H}|}$ represent
the weights and bias of the last fully-connected layer. An image $\mathbf{x}_i$
is mapped to its feature representation $\mathbf{h}_i$ by all but one layers of
the network $f$ and the predicted label $\hat{f}_{\mathcal{H}}(\mathbf{x}_i)$ is
determined by the index of the largest element in the vector
$\mathbf{W}\mathbf{h}_i + \mathbf{b}$, as follows:
\begin{equation}
  \hat{f}_{\mathcal{H}}(\mathbf{x}_i) = \arg\max_{c'} \langle \mathbf{w}_{c'}, \mathbf{h}_i \rangle + b_{c'},
\end{equation}
where $\mathbf{w}_{c'}$ is the $c'$-th row of $\mathbf{W}$.

The global mean is defined as $\bm{\mu}_G \triangleq
\text{Ave}_{i,c}\{\mathbf{h}_{i,c}\}$, and the train class means are defined as:
\begin{equation}
\label{eq:class_means}
\bm{\mu}_c \triangleq \text{Ave}_{i}\{\mathbf{h}_{i,c}\}, \quad c
= 1, \dots, |\mathcal{H}|,
\end{equation}
where \(\text{Ave}\) is the averaging operator.

Additionally, the between-class covariance, $\boldsymbol{\Sigma_B} \in \mathbb{R}^{p
  \times p}$ is defined as:

\begin{equation}
  \boldsymbol{\Sigma_B} \triangleq \text{Ave}_{c} \left\{ (\bm{\mu}_c - \bm{\mu}_G)(\bm{\mu}_c - \bm{\mu}_G)^\top \right\},
\end{equation}

and the within-class covariance, $\boldsymbol{\Sigma_W} \in \mathbb{R}^{p \times
  p}$ is defined as:

\begin{equation}
  \mathbf{\Sigma_W} \triangleq \text{Ave}_{i,c} \left\{ (\mathbf{h}_{i,c} - \bm{\mu}_c)(\mathbf{h}_{i,c} - \bm{\mu}_c)^\top \right\}.
\end{equation}

NC manifests through four key observations as described by \cite{papyan2020prevalence}.

\textbf{(NC1)} Variability collapse: $\boldsymbol{\Sigma_W} \rightarrow \mathbf{0}$.

Following Papyan et al. (2020), in the Figure 5 of the paper, we plot $\text{tr}
\left( \Sigma_W \Sigma_B^\dagger / C \right)$, where tr denotes the trace
operation and $[ \cdot ]^\dagger$ is the Moore–Penrose pseudoinverse.

\textbf {(NC2)} Convergence to simplex ETF:
\begin{equation}
  |\|\bm{\mu}_c - \bm{\mu}_G\|_2 - \|\bm{\mu}_{c'} - \bm{\mu}_G\|_2| \to 0 \quad \forall \, c, c'
\end{equation}

\begin{equation}
  \langle \tilde{\bm{\mu}}_c, \tilde{\bm{\mu}}_{c'} \rangle \to \frac{C}{C-1} \delta_{c,c'} - \frac{1}{C-1} \quad \forall \, c, c',
\end{equation}
where $\tilde{\bm{\mu}}_c = \frac{(\bm{\mu}_c - \bm{\mu}_G)}{\|\bm{\mu}_c -
  \bm{\mu}_G\|_2}$ are the renormalized class means.

Following the original paper, we plot the following metrics in Section~\ref{sec:nc_addings}:
\begin{enumerate}
\item Norms equality level measured by two ratios: a) the standard deviation to
  the average length of class means and b) the standard deviation of the length
  of the rows of the weight matrix to their average length:
 \begin{equation}
   \beta_{\mu} = \text{Std}_c \left( \|\bm{\mu}_c - \bm{\mu}_G\|_2 \right) / \text{Avg}_c
   \left( \|\bm{\mu}_c - \bm{\mu}_G\|_2 \right),
  \end{equation}
\begin{equation}
  \beta_{w} =\text{Std}_c \left(
     \|\mathbf{w}_c\|_2 \right) / \text{Avg}_c \left( \|\mathbf{w}_c\|_2
  \right),
\end{equation}
where $\mathbf{w}_c$ is the c-th row of $\mathbf{W}$ (classifier of the c-th
class).
\item Angles equality level measured as the standard deviation of cosines of the
  angles between class means ($\alpha_{\mu}$) and the standard deviation
of cosines of angles between rows of $\mathbf{W}$ ($\alpha_{w}$).
\end{enumerate}

\textbf{(NC3)} \text{ Convergence to self-duality:}

\begin{equation}
  \label{eq:self-duality}
\left\| \frac{\mathbf{W}^\top}{\|\mathbf{W}\|_F} - \frac{\dot{\mathbf{M}}}{\|\dot{\mathbf{M}}\|_F} \right\|_F \to 0,
\end{equation}
where $\dot{\mathbf{M}} = [\bm{\mu}_c - \bm{\mu}_G, c = 1, \dots, C] \in
\mathbb{R}^{p \times C}$ is the matrix obtained by stacking the class means into
the columns of a matrix. Here, $\delta_{c, c'}$ is the Kronecker delta symbol.

We report value, estimated by equation~(\ref{eq:self-duality}), in Figure 5.

\textbf{(NC4)} Simplification to Nearest Class Centroid (NCC):
\begin{equation}
  \arg\max_{c'} \langle \mathbf{w}_{c'}, \mathbf{h} \rangle + b_{c'} \to \arg\min_{c'} \|\mathbf{h} - \bm{\mu}_{c'}\|_2.
\end{equation}
We report the proportion of mismatch between two ways of label estimation in
Figure 5.

\textbf{Hypernym Label Spaces}. To estimate NC in the hypernym label space
$\mathcal{S}$, we compute the mean of each superclass $s \in \mathcal{S}$:

\begin{equation}
  \bm{\mu}_s \triangleq \text{Ave}_{c \in s}\{\bm{\mu}_c\}.
\end{equation}

And we assume that the superclass weight matrix $\mathbf{W}_s \in
\mathbb{R}^{|\mathcal{S}_n| \times p}$ consists of $|\mathcal{S}_n|$ rows, where
each row $\mathbf{w}_s$ is obtained by averaging the rows of $\mathbf{W}$:

\begin{equation}
  \mathbf{w}_{s} \triangleq \text{Ave}_{c \in s}\{\mathbf{w}_c\},
\end{equation}

and similarly for the bias vector $b_s \triangleq \text{Ave}_{c \in s}\{b_c\}$.

We can then use $\{\bm{\mu}_s\}, \mathbf{W}_s$, and $\mathbf{b}_s = [b_s]$
instead of $\{\bm{\mu}_s\}, \mathbf{W}$, and $\mathbf{b}$ as defined above, to
estimate Neural Collapse in the hypernym spaces $\mathcal{S}$.

\section{EXPERIMENTAL DETAILS}
\label{appendix_exp_details}

In this section, we provide additional details on the experimental setup for the
experiments discussed in Section~4.

\subsection{Architectures and Training Parameters}

The primary conclusions are based on experiments with ResNet; however, we also
include results for ViT and MobileNet V3-L 1.0 to demonstrate that the findings
generalize beyond convolutional or parameter inefficient architectures. Table
\ref{tab_net_sizes} summarizes key parameters of the neural networks used in
this study.

\begin{table}[h]
  \caption{\label{tab_net_sizes}Networks used in the experiments.}
  \centering
  \begin{tabular}{llrr}
    \textbf{Name} & \textbf{Basic Block} & \textbf{\#Params (M)} & \textbf{Top-1 Accuracy (\%)} \\
    \hline
    ResNet-18          & Convolution & 12.0 & 69.8 \\
    ResNet-50          & Convolution & 25.0 & 79.0 \\
    ResNet-152         & Convolution & 60.2 & 65.0 \\
    ViT-B/16           & Attention   & 86.0 & 75.6 \\
    MobileNet V3-L 1.0 & Convolution & 5.4  & 75.6 \\
  \end{tabular}
\end{table}

All architectures were trained using 224 $\times$ 224 images. For ResNet-18,
ResNet-50 and ViT-B/16 the following hyperparameters we used:
\begin{itemize}
    \item 200 epochs;
    \item Cosine learning rate schedule;
    \item SGD optimizer with an initial learning rate of 0.05;
    \item AutoAugment \citep{cubuk2018autoaugment};
    \item Loss function with Jensen-Shannon divergence.
\end{itemize}

MobileNet was trained for 600 epochs using RMSprop, with noise added to the
learning rate, a batch size of 512, warmup, and Exponential Moving Average
(EMA).

ViT training resulted in relatively low accuracy. Nevertheless, the presence of
hypernym bias under suboptimal training conditions supports the hypothesis that
the phenomenon is general.

ResNet-152 was trained in neural collapse settings \citep{papyan2020prevalence}:
\begin{itemize}
  \item No augmentation;
  \item 600 images per class in the train set;
  \item Batch size of 256;
  \item 350 epochs;
  \item SGD optimizer with initial learning rate 0.1;
  \item Learning rate is annealed by a factor of 10 at 1/2 and 3/4 of epochs.
\end{itemize}

These hyperparameters achieve nearly 100\% accuracy on training set, which is
necessary for NC to manifest, although validation accuracy is low in this case.

\subsection{Constructing Hypernym Label Spaces}

\textbf{High-Level Hypernyms}. At the top level of the WordNet tree, there
are 9 synsets:
\begin{enumerate}
    \item Plants (2 classes in ImageNet),
    \item Geological formations (10 classes),
    \item Natural objects (16 classes),
    \item Sports (0 classes),
    \item Fungi (7 classes),
    \item People (3 classes),
    \item Miscellaneous (42 classes),
    \item Artifacts (522 classes),
    \item Animals (398 classes).
\end{enumerate}

Since the synsets have an uneven distribution of classes, we used the following
groupings to form a top-level hypernym space, $\mathcal{S}_3$:
\begin{enumerate}
    \item \{1-7\} (miscellaneous, 80 classes),
    \item \{8\} (artifacts, 522 classes),
    \item \{9\} (animals, 398 classes).
\end{enumerate}

\textbf{Low-Level Hypernym spaces}. The low-level hypernym superclasses were
created according WordNet tree. Similar to the 3 top-level hypernym
superclasses, we aimed to maintain approximately equal class saturation in
ImageNet classes, and some superclasses were formed by grouping several synonym
sets at the same hierarchical level. For instance, to obtain 7 superclasses:
\begin{itemize}
    \item The ``miscellaneous'' superclass remained unchanged.
    \item The ``animals'' superclass was divided into 3 new superclasses based on
      the lower levels of the WordNet hierarchy:
    \begin{enumerate}
    \item Chordates (207 classes),
    \item Domestic animals (123 classes),
    \item Others (including invertebrates (61 classes) and game birds (7
      classes)).
    \end{enumerate}
\end{itemize}

Similarly, the ``artifacts'' superclass was divided into the following three
superclasses:
\begin{enumerate}
\item Instrumentality (358 classes),
\item Covering (85 classes),
\item Others (including categories like ``commodity'' and ``decoration'',
  totaling 79 classes).
\end{enumerate}

For a more granular categorization, we divided the ``animals'' superclass into
18 new superclasses, such as ``domestic cat'', ``reptile'', and ``terrier'',
each containing an average of 22 ImageNet classes.

The grouping procedure was performed only once in all cases. This was done to
maintain consistency in the sizes of the hypernym sets, ensuring that the
reported metrics are comparable. Highly imbalanced superclasses tend to yield
higher accuracy regardless of their number, as can be seen from equations in
Section \ref{appendix:theoretical_acc}.

\section{ADDITIONAL EXPERIMENTS}
\label{appendix_add_experiments}
\subsection{Greedy Hypernym Classification}
\subsubsection{Confusion Matrix Evolution}

\begin{figure}[h]
  \centering
  \includegraphics[width=0.99\linewidth]{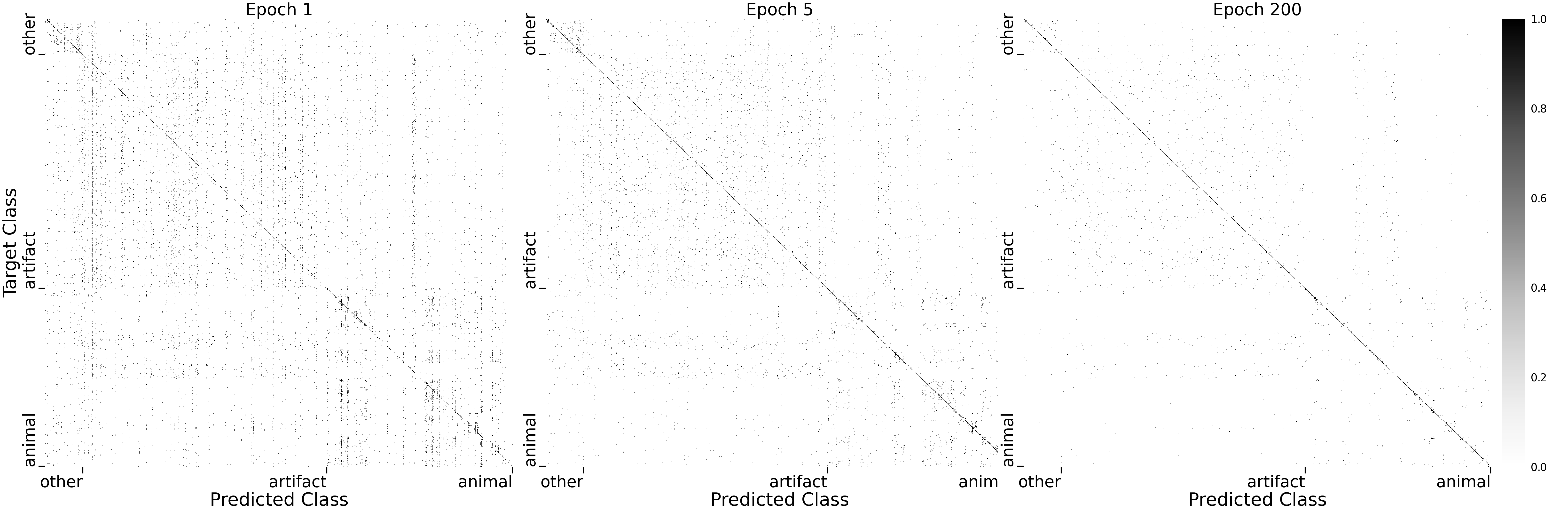}
  \caption{\label{fig_cm}Confusion matrix of ResNet-50 during training. In early
    stages of training block diagonal structure is clearly visible.}
\end{figure}

A straightforward way to observe the manifestation of hypernym bias is by
plotting a confusion matrix with class labels arranged according to the
structure of the WordNet hierarchy. In this case, the order of classes was
determined by performing a depth-first traversal of the tree, ensuring that
sibling classes appear consecutively in the confusion matrix. We visualize the
confusion matrices at the 1st, 5th, and 200th epochs of training ResNet-50 in
\figurename~\ref{fig_cm}.

In the early stages, the block-diagonal structure is apparent, although the
WordNet tree only partially mirrors the misclassification patterns. This can be
observed in the structural discontinuities within the blocks.

\subsubsection{Residual Error}
\label{appendix_resid_error}

To provide a clearer understanding of the training dynamics of ResNet-50, as
shown in Figures 1 and 3, we introduce the residual relative error metric, \(
E_R \), defined as:

\begin{equation}
  \label{eq_resid_error}
  E_R(\mathcal{X},t) = \frac{100 - A(\mathcal{X}, t)}{100 - A(\mathcal{X}, T)} -1,
\end{equation}
where $A(\mathcal{X}, t)$ is the accuracy at epoch $t$, and $A(\mathcal{X}, T)$
is the final accuracy after $T$ epochs.

\begin{figure}[h]
  \centering
  \includegraphics[width=12cm]{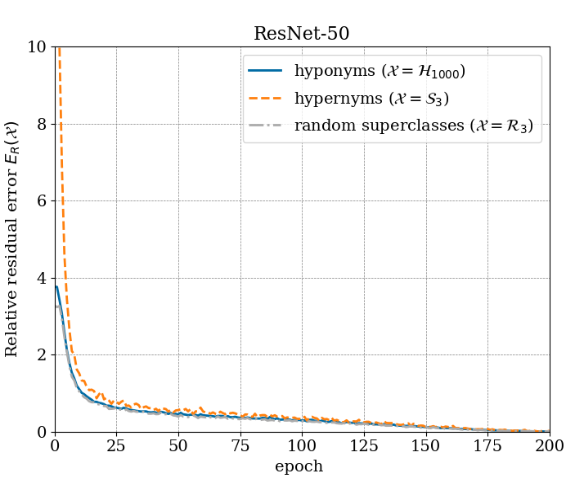}
  \caption{\label{fig_figure2_appendix}Residual relative error  over the course of
    ResNet-50 training. }
\end{figure}

Figure \ref{fig_figure2_appendix} illustrates the residual relative error over
the course of ResNet-50 training. The graph shows that the error is
significantly higher for hypernym recognition during the first 10 to 20 epochs.
After this initial phase, the residual relative error becomes similar across all
three label spaces, indicating that the majority of the improvement in hypernym
recognition accuracy occurs early in the training process.

\subsubsection{Different Hypernym Spaces}
\label{appendix:animals_groups}

\textbf{Animal Hypernym Space}. We examined the recognition of images within the
``animals'' hypernym label space using ResNet-50. This analysis is motivated by
the well-known hierarchical structure of species, which resembles a tree-like
hierarchy. Although the WordNet tree has known limitations, this structure is
likely to provide a more accurate description for this portion of the ImageNet
dataset. The results are presented in Table \ref{tab_superclasses_animal}.

\begin{table}[h]
  \caption{\label{tab_superclasses_animal}Number of epochs required by ResNet-50
    to achieve 95\% relative accuracy in different labels spaces with varying
    sizes (animal images only).}
\centering
\begin{tabular}{rrr}
  $\boldsymbol{n}$ & $\boldsymbol{\mathcal{S}_n}$ & $\boldsymbol{\mathcal{R}_n}$ \\
  \hline
  2  & 6  & 46 \\
  3  & 12 & 96 \\
  4  & 13 & 105\\
  7  & 20 & 109\\
  8  & 25 & 114\\
  18 & 48 & 114\\
\end{tabular}
\end{table}

Comparing the results in Table \ref{tab_superclasses_animal} with those for the
entire set of classes (Figure~4), we observe that for a random division into 7
superclasses, 95\% accuracy is achieved in 109 epochs for both cases. However,
when utilizing the animal hierarchy, training converges faster, reaching the
target in 20 epochs compared to 25.

\textbf{Unbalanced hypernym space}. To demonstrate that intuitive \textit{manual}
grouping of synsets inside the same hypernym level of the tree does not affect
the conclusions, we also evaluated superclasses derived directly from one of the
levels of the WordNet hierarchy without additional grouping for class balance.
This resulted in 72 unbalanced superclasses, with the corresponding results
presented in Table \ref{tab_superclasses_72}.

\begin{table}[h]
  \caption{Number of epochs to achieve 95\% recognition accuracy using 72
    superclasses} \label{tab_superclasses_72}
\centering
\begin{tabular}{lrrr}
  \textbf{Architecture} & $\boldsymbol{\mathcal{R}_{72}}$ & $\boldsymbol{\mathcal{S}_{72}}$ \\
  \hline
  ResNet-50 & 114 & 31\\
  ResNet-18 & 88  & 31\\
\end{tabular}
\end{table}

\subsubsection{Other datasets}
\label{appendix_other_sets}

We have conducted additional experiments to provide further evidence of hypernym
bias existence across different datasets and domains.

\textbf{CIFAR-100}. CIFAR-100 comprises 60 000 32x32 images distributed across
100 classes, which are further grouped into 20 superclasses. Let
$\mathcal{S}_{20}$ represent the hypernym label space (superclasses) and
$\mathcal{R}_{20}$ denote a random label space isomorphic to $\mathcal{S}_{20}$.

We trained cifar-adapted ResNet-18 with augmentation and cosine annealing of
learning rate and achieved 77\% validation accuracy in the hyponym label space.

We employed the greedy classifier model described in Section 3.1 to estimate
relative accuracy in hypernym and random label spaces. The results demonstrate
an accelerated growth of $A_R(\mathcal{S}_{20})$ compared to
$A_R(\mathcal{R}_{20})$, which aligns with our expectations. The manifestation
of hypernym bias is evident but less pronounced than in Figure 1a (ImageNet).
This discrepancy is fully explainable:

\begin{enumerate}
\item The hierarchical structure of CIFAR-100 is less well-defined compared to
  WordNet, with loosely distinct superclasses such as vehicles\_1 and vehicles\_2.
\item As noted in Section 4.1, larger hypernyms tend to exhibit a more pronounced
  bias. In CIFAR-100, each superclass comprises only 5 labels.
\end{enumerate}

\begin{figure}[h]
  \centering
  \includegraphics[width=.95\linewidth]{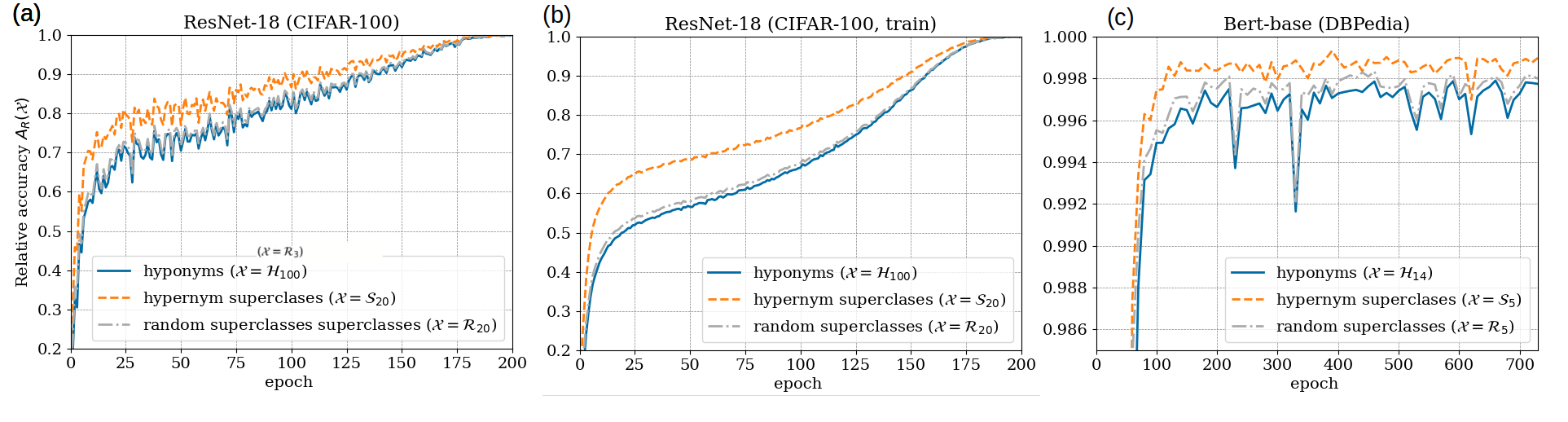}
  \caption{\label{fig_cifar_dbpedia} Generalization of hypernym bias. Relative
    accuracy in different labelspaces for CIFAR-10 (a,b) and DBPedia (c)
    classifiers during course of training. Graph for DBPedia shows only first
    750 training iterations}
\end{figure}

In this setup, the difference in dynamics between hypernyms and random
superclasses is most evident on the training set, where the estimates are less
noisy. This is particularly noticeable when $A_R(\mathcal{X}) \approx 0.6$ as
shown in \figurename~\ref{fig_cifar_dbpedia}.

We also observe a consistent manifestation of hypernym bias across a variety of
training parameters on CIFAR-100.

\textbf{DBPedia}. DBPedia is a text classification dataset containing 630,000
samples. It includes 14 classes, which we grouped into 5 hypernyms
($\mathcal{S}_{5}$) based on the first level of the DBPedia ontology as shown in
\tablename~\ref{tab:superclasses_and_classes}.

\begin{table}[h!]
  \centering
  \caption{Superclasses structure according to DBPedia ontology}
  \begin{tabular}{lp{0.6\textwidth}}
    \textbf{Superclass} & \textbf{Classes} \\
    \hline
    Eukaryote & Artist, Athlete, OfficeHolder, Animal, Plant \\
    Organization & Company, EducationalInstitution \\
    Place & NaturalPlace, Village \\
    Work & Album, Film, WrittenWork \\
    Others & MeanOfTransportation, Building \\
  \end{tabular}
  \label{tab:superclasses_and_classes}
\end{table}

We fine-tuned a pretrained BERT-base model on DBPedia for 3 epochs using a batch
size of 32 and a learning rate of $2 \cdot 10^{-5}$. The test errors were as
follows: 0.67\% for hyponym labels, 0.48\% for random superclass labels, and
0.37\% for hypernym labels. The relative accuracy curves demonstrate that the
difference between $A(\mathcal{R}_5)$ and $A(\mathcal{S}_5)$ emerges during the early
iterations, as predicted. Hypernym bias is evident.


Thus, we demonstrated the generalizability of our approach to two domains
(images and text)

\subsection{Neural collapse}
\label{sec:nc_addings}
\begin{figure}[h]
  \centering
  \includegraphics[width=.8\linewidth]{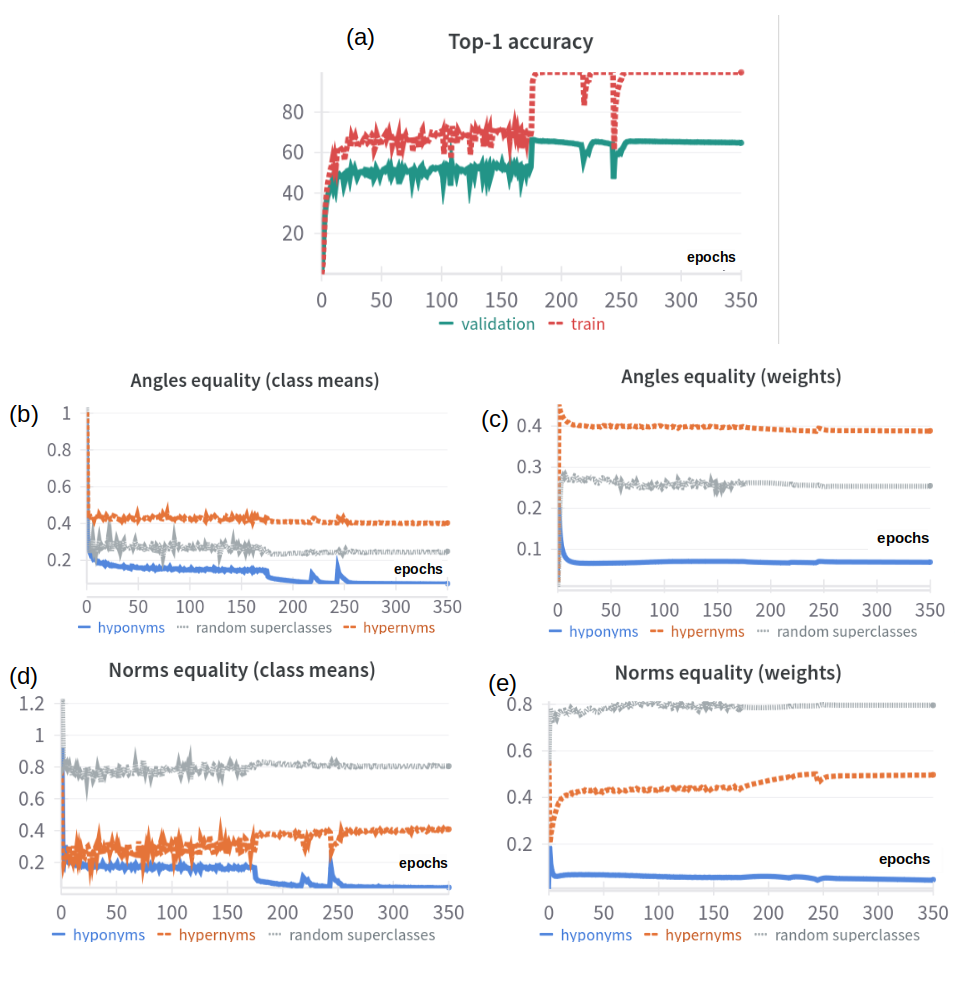}
  \caption{\label{fig_nc_curves_appendix} Train and validation accuracy during
    training in NC settings (a), level of angle equality between class means
    $\alpha_{\mu}$ (b)
    and rows of weights matrix $\alpha_{w}$ (c), level of norm equality of class
    means $\beta_{\mu}$ (d)
    and rows of weights matrix $\beta_{w}$ (e).}
\end{figure}

\textbf{ETF convergence estimation}. Figure 5 shows kthe curves for NC1, NC3, and
NC4 in the label spaces of top-level hypernyms \( \mathcal{S}_3 \), random
superclasses \( \mathcal{R}_3 \), and hyponyms \( \mathcal{H}_{1000} \). Here,
in Figure \ref{fig_nc_curves_appendix}, we provide additional graphs that
illustrate the level of convergence to the equiangular tight frame (ETF), as
computed according to Appendix~\ref{sec:nc_theory}.

As can be seen from the figure we do not observe convergence to ETF in the hypernym label space.

\begin{figure}[h]
  \centering
  \includegraphics[width=.7\linewidth]{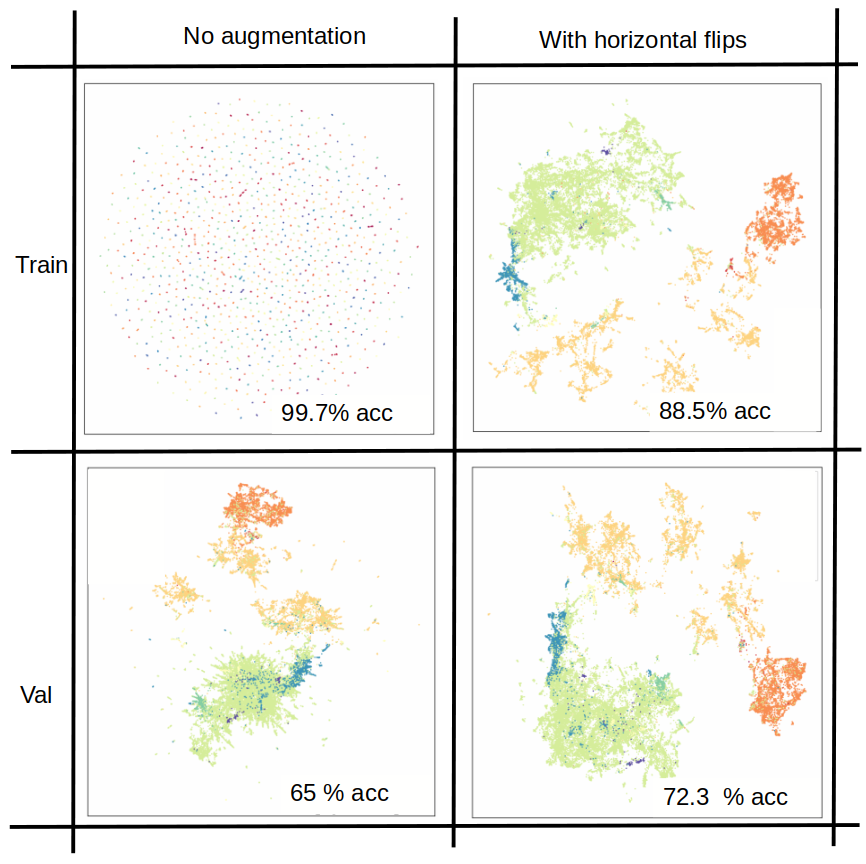}
  \caption{\label{fig_imagenet_nc_compare_end}UMAP embeddings of penultimate
    features for 50 training and 50 validation examples per class in the end of
    training. The color indicates the WordNet distance to the anchor class.}
\end{figure}

\textbf{UMAP visualization}. In Figure~\ref{fig_imagenet_nc_compare_end}, we show
UMAP embeddings of the penultimate features of two versions of trained
ResNet-152: one trained with and one trained without horizontal flip
augmentation.

\begin{figure}[h]
  \centering
    \includegraphics[width=.9\linewidth]{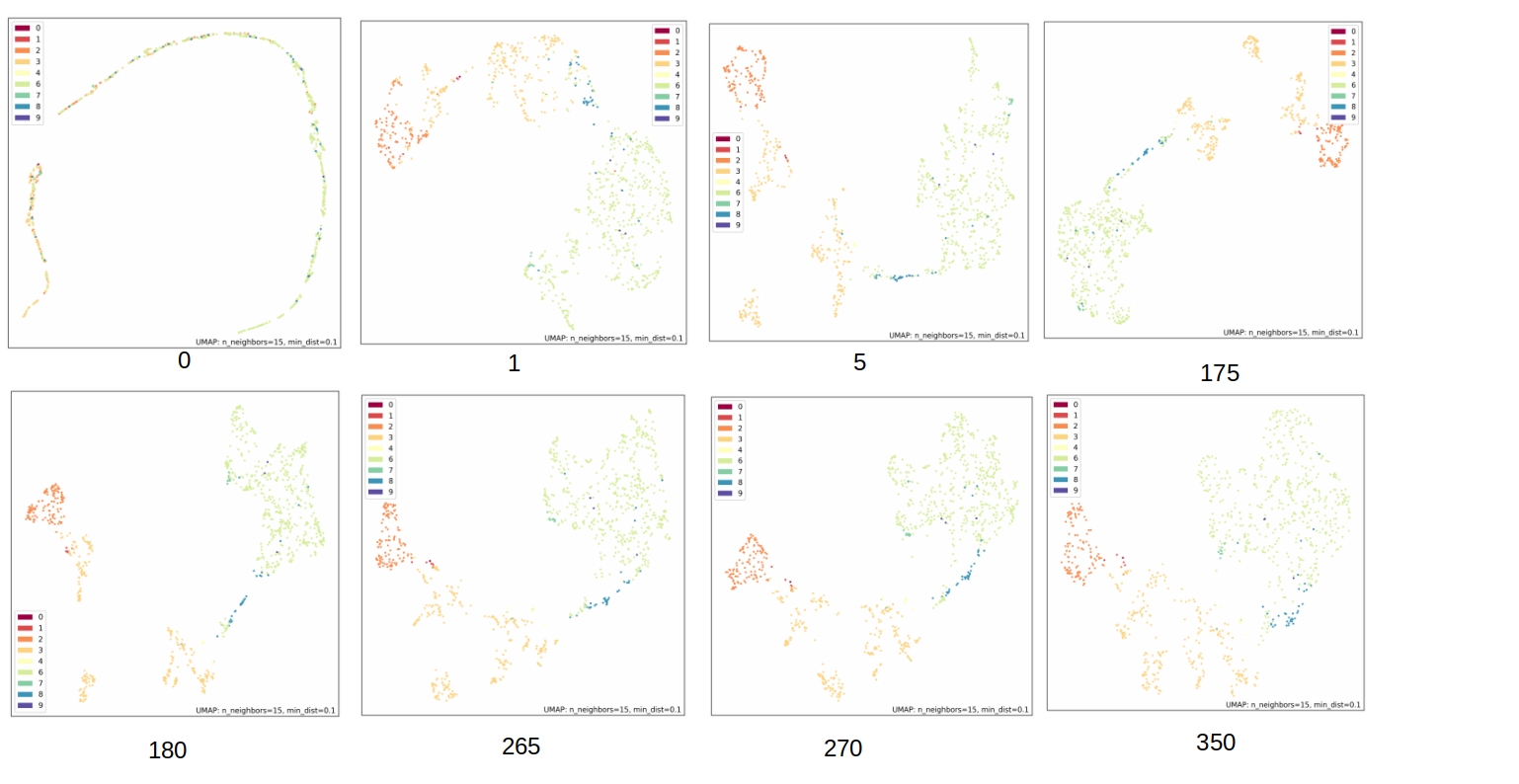}
    \caption{\label{fig_imagenet_nc_cs} UMAP embeddings of the class means of
      the penultimate features on different epochs. The color indicates the
      WordNet distance to the anchor class.}
\end{figure}

As shown in Figure~\ref{fig_imagenet_nc_compare_end}, data augmentation prevents
the neural network from reaching 100\% accuracy, and Neural Collapse (NC) is not
observed. Similarly, NC is absent on the validation set. The graphs in Figure 5
of and Figure \ref{fig_nc_curves_appendix} similarly to those in (Papyan et al.,
2020), display only moderate convergence towards neural collapse in ImageNet.
However, NC is evident in UMAP embeddings shown in Figure 1, suggesting complete
removal of hyponymy information from penultimate features. Yet, when we plot the
UMAP embeddings of the class centers, the hypernym structure remains preserved
until the final epoch, although information gradually decays (see
Figure~\ref{fig_imagenet_nc_cs}).

This is likely due to averaging, which effectively reduces noise. We anticipate
that training a larger network for a longer period will result in full NC on
ImageNet.

\subsection{Manifold Distances}

Here we briefly compare the method of defining class distances based on mutual
covers in feature space, as described in Section 3.2, with other approaches.
Interestingly, these simpler methods also exhibit high cophenetic correlation
coefficient (CCC) between the distances in feature space and hyponym distance in
the WordNet graph.

Let $\mathbf{h}^l_{i,c}$ be ResNet-50 feature of layer $l$ in response to an
image with hyponym label $c$ \footnote{For layers below pre-logit layer before
  estimating distances we applied global max-pooling to the feature maps}. We
use UMAP with default parameters to create 10-dimensional embeddings
$\boldsymbol{\phi}^l_{i,c}$ of the features, these embeddings are used further.

\textbf{Class means distance}. We can construct the class distance matrix $D_F$
using distances between class means $d_f(c,c')=\| \boldsymbol{\mu_c} -
\boldsymbol{\mu_{c'}} \|_2$, where class means $\bf{\mu_c}$ is average
embeddings of class $c$ (similarly to equation~(\ref{eq:class_means})).

\textbf{Correlation from direct comparison of examples}. We can omit
materializing class distances by computing correlation coefficient directly
between individual distances $d_f(u^{c_i},v^{c_j})$ between examples of
different classes in the feature (embedding) space and distances $d_w(c_i, c_j)$
in the WordNet graph.

Results of all methods are reported in \tablename~\ref{tab:ccc_appendix}.

\begin{table}[h]
  \caption{\label{tab:ccc_appendix}Cophenetic correlation coefficient of WordNet
    graph distances and distances between features of different labels
    (ResNet-50)}
  \begin{center}
    \begin{tabular}{lccc}
      \textbf{Layer name}            & \textbf{Direct comparison}    & \textbf{Class Means } & \textbf{Mutual Cover } \\ \hline
      Layer1 (max pool)     & 0.06                  & 0.26                  & \textbf{0.31} \\
      Layer2 (max pool)     & 0.20                  & 0.44                  & \textbf{0.44} \\
      Layer3 (max pool)     & 0.49                  & \textbf{0.62}                  & 0.56 \\
      Pre-logits            & 0.58                  & \textbf{0.61}                  & 0.59 \\
      Logits                & 0.59                  & \textbf{0.62}                  & 0.60 \\
      \hline
      Average               & 0.32                  & 0.44                  & \textbf{0.45} \\
    \end{tabular}
  \end{center}
\end{table}

From the table, method based on mutual cover provides higher CCC in bottom
layers, while class means distances preserve more information about hypernym
relation in top layers. Mutual cover distances, though, provide CCC that
gradually increase from bottom to top layers.

\subsection{Linear Probes}
\label{appendix:probes}

We experimented with two methods of training the linear probes:

\begin{enumerate}
\item \textbf{Additional fully-connected layers.} This technique
  was proposed by \cite{alain2016understanding}. The additional layers are
  trained simultaneously with the entire network, but gradients from the probes
  do not propagate into the network itself.

\item \textbf{Few-shot learning.} In this approach, features from several
  examples per class are collected during a specific phase of training and used
  to train a linear model. This probing method has been employed in previous
  studies, such as \citep{Dosovitskiy2020,raghu2021vision}. We applied two types
  of linear learners:
\begin{itemize}
\item An analytical solution using the least squares method with regularization
  as proposed by \cite{raghu2021vision}.
\item Logistic regression.
\end{itemize}
\end{enumerate}

The advantage of using additional layers trained simultaneously with the network
is that the entire training set can be fully utilized, though the adaptation is
influenced by the continuously changing weights of the neural network. In
contrast, few-shot learning requires more memory but provides a solution
tailored to the specific state of the network.

In all cases, we trained the probes in the hyponym label space \(
\mathcal{H}_{1000} \) and evaluated accuracy in both the hypernym label space \(
\mathcal{S}_3 \) and the hyponym label space \( \mathcal{H}_{1000} \). Training
in \(\mathcal{H}_{1000} \) provides a regularization effect and is more
effective in hypernym labels too, as shown in Table
\ref{tab_label_spaces_probes}.

\begin{table}[ht]
  \centering
  \caption{\label{tab_label_spaces_probes} Hypernym linear probes classification
    accuracy $A(\mathcal{S}_3)$ using different label spaces for training.
    Results are for the ResNet-18 layers and the logistic regression probes}
  \begin{tabular}{ccc}
    \textbf{Layer} & \textbf{Hypernym} $\mathbf{(\mathcal{S}_3)}$ & \textbf{Hyponym} $\mathbf{(\mathcal{H}_{1000})}$ \\
    \hline
    layer1 & 70,4 & \textbf{73,2} \\
    layer2 & 77,9 & \textbf{79,6} \\
    layer3 & 87,3 & \textbf{87,7} \\
    layer4 & 93,4 & \textbf{95,5} \\
    logits & 92,6 & \textbf{95,8} \\
  \end{tabular}
\end{table}

\subsubsection{Additional Layers}

The linear layers were trained for ResNet-18 using the same optimizer and
parameters as the entire network, but with cross-entropy loss as the objective
function for all intermediate layers, without any additional loss components.

We followed the recommendations of \cite{alain2016understanding}:
\begin{itemize}
\item A linear layer is added at the end of each convolutional block.
\item The weights of the linear layer are randomly initialized.
\item To reduce the number of parameters, the feature maps are downsampled in
  both height and width using average pooling by a specified factor.
\item Evaluation is conducted on the validation set.
\end{itemize}

We conducted experiments using feature maps at various resolutions using
different average pulling windows.

\subsubsection{Few-Shot Linear Learner}

For few-shot training, we use 10 examples per class sampled from a portion of
the validation set, and evaluate the classifier on the remaining part of the
validation set.

Initially, we followed the approach described in \citep{raghu2021vision}, which
employs an analytical solution. Additional investigations, described further,
revealed that the analytical solution performed significantly worse in the early
layers of the neural network compared to trainable linear layers. We found that
iterative logistic regression yielded slightly more robust results, which are
reported in the paper, though it is still inferior to trained layers.

\textbf{ResNet}. For ResNets we used features obtained similarly to trained
probes with medium size average pooling window.

\textbf{ViT}. For ViT, to connect probes we tested the averaged token embedding
as suggested by \cite{chen2020generative} and classification token embedding
only as done in \citep{raghu2021vision}. Classification token worked
significantly better and we report results with the usage of it only. The
transformer-decoder processes a sequence of discrete input tokens, \(x_1, \dots,
x_n\), and generates a \(d\)-dimensional embedding for each position. The
decoder consists of a stack of \(L\) blocks, with the \(l\)-th block producing
an intermediate embedding, \(h^1, \dots, h^l\), each of dimension \(d\).

The ViT block updates the input tensor \(h^l\) as follows:
\[
  \begin{aligned}
    n^l &= \text{layer\_norm}(h^l), \\
    a^l &= h^l + \text{multihead\_attention}(n^l), \\
    h^{l+1} &= a^l + \text{mlp}(\text{layer\_norm}(a^l)).
  \end{aligned}
\]

The probe is taken based on the first normalized classification token embedding
$n_0^l$ at each layer $l$, where 0 is the index of classification token.

\subsubsection{Results}

\begin{figure}[h]
\centering
\includegraphics[width=9cm]{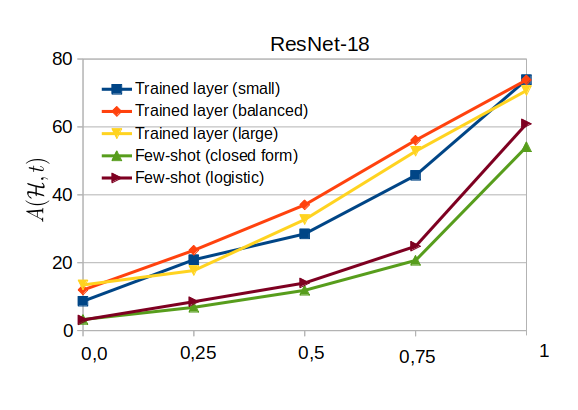}
\caption{\label{fig_resnet18_probes_types}Comparison of probes types for fully
  trained ResNet-18. Small, balance, and large trainable probes were obtained
  using features with large, medium and small pooling windows, respectively. The
  horizontal axis represents the normalized layer depth.}
\end{figure}

\textbf{Comparison}. For ResNet-18, training probes provide significantly better
results than few-shot learning (Figure~\ref{fig_resnet18_probes_types}).
Logistic probes are slightly better than those based on analytical solutions.

However, we have observed that trainable layers doesn't change relative accuracy
of probing and the same conclusions can be made. Therefore, in the paper we
present results for logistic regression probes only, but we verify conclusions
for several architectures.

\begin{figure}[h]
\centering
\includegraphics[width=14cm]{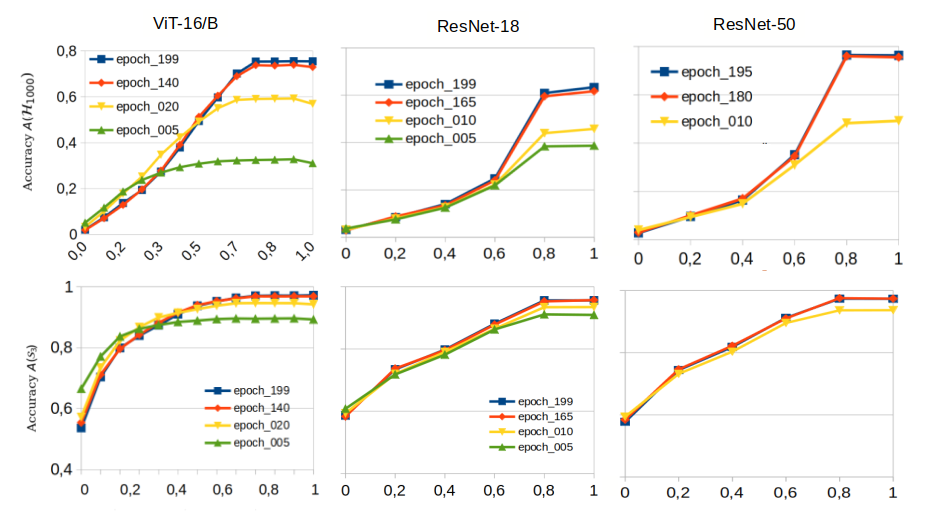}
\caption{\label{fig_logistic_appendix}Classification accuracy based on linear
  probes for different architectures in different phases of training. The
  horizontal axis of each graph represents the normalized layer depth. The top
  row corresponds to the hyponym label space \( \mathcal{H}_{1000} \), while the
  bottom row corresponds to the hypernym label space \( \mathcal{S}_3 \).}
\end{figure}

\textbf{Probes for ViT and ResNet-18.} Figure 6 present results for few-shot
linear probes applied to ResNet-50. Here, in Figure \ref{fig_logistic_appendix},
we extend the analysis to include additional experiments with ViT-B/16 and
ResNet-18.

The following conclusions can be drawn from the graphs in Figure
\ref{fig_logistic_appendix}:

\begin{enumerate}
\item Classification accuracy improves as the network progresses towards the top
  layer, for both hyponym and hypernym classification. This trend is observed
  across all architectures.

\item During the early stages of ViT training, classification accuracy based on
  the initial layers is higher compared to the end of training. This pattern is
  evident for both the hyponym and hypernym label spaces and can be attributed
  to the removal of excess information from the classification token as training
  progresses.

\item In all experiments, hypernym classification accuracy using intermediate
  layers is consistently lower than when using the top layer. Therefore, we
  cannot conclude that the early layers of the neural network contain sufficient
  information for accurate hypernym recognition.
\end{enumerate}

\subsection{Image Frequency Analysis}
\label{appendix_freq_analysis}
\begin{figure}[h]
  \centering \includegraphics[width=14cm]{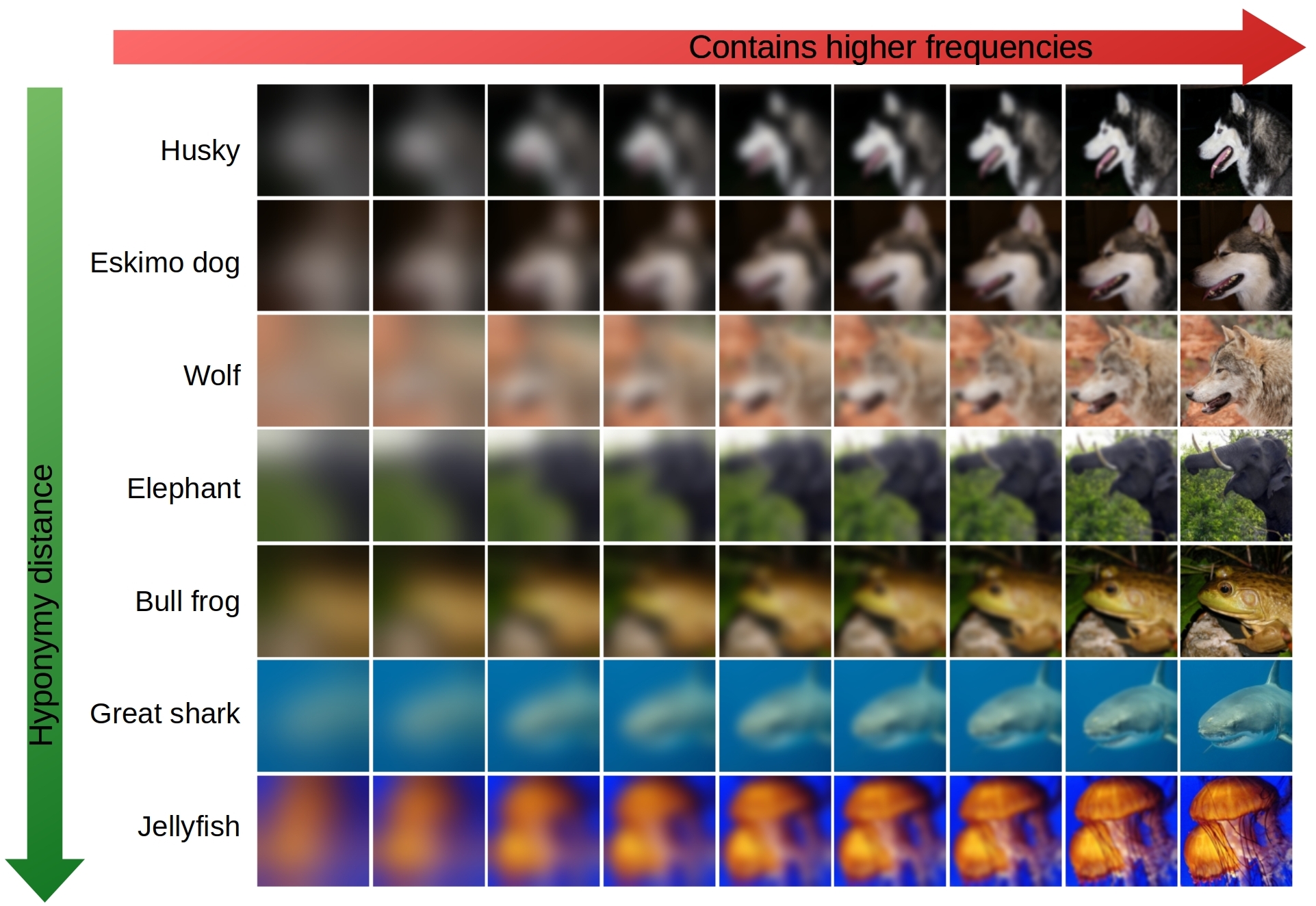}
  \caption{\label{fig_blur_demo} Examples of images from classes with different
    hyponymy distances and varying amounts of high-frequency content. Each
    column applies the same low-pass filter parameters, producing a consistent
    level of detail across all classes, while rows represent different
    categories with increasing hyponymy distance from the class presented in the
    top row.}
\end{figure}

One can suggest that recognizing hypernyms is possible based on low image
frequencies, which are learned in first phase of training, while hyponyms
require high frequencies that are learned later
(\figurename~\ref{fig_blur_demo}). This explanation through Frequency Principle
seems intuitive and lead to interesting interpretations of the fact that infants
perceive world in low frequencies. However our experiments did't reveal this
relation.

We removed high frequencies from the images using Gaussian kernel of different
size. The focus was on the accuracy of hypernym recognition at the beginning and
at the end of training when using a significant level of blurring. For this, we
used the 10th and 195th epochs of ResNet-50 training and the 5th and 140th
epochs of ViT-B/16.

\begin{figure}[h]
\centering
\includegraphics[width=12cm]{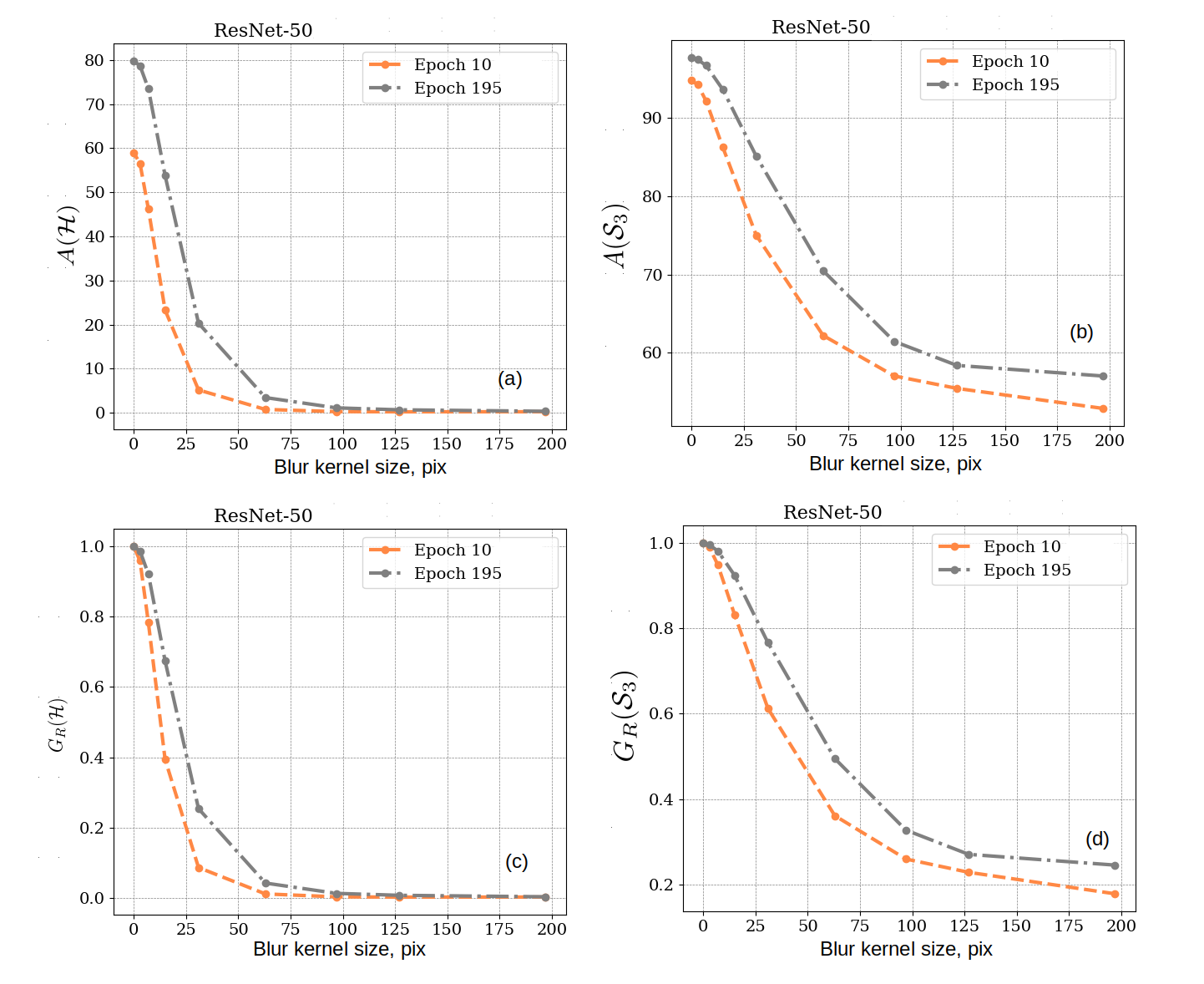}
\caption{\label{fig_blur_resnet} Accuracy (a,b) and relative accuracy gain (c,d)
  of ResNet-50 in hyponym label space (a,c) and hypernym label space (b,d) with
  the respect to size of the blur kernel. $G_R$ is computed with respect to
  accuracy on undistorted images for the same checkpoint.}
\end{figure}

\begin{figure}[h]
  \centering
  \includegraphics[width=12cm]{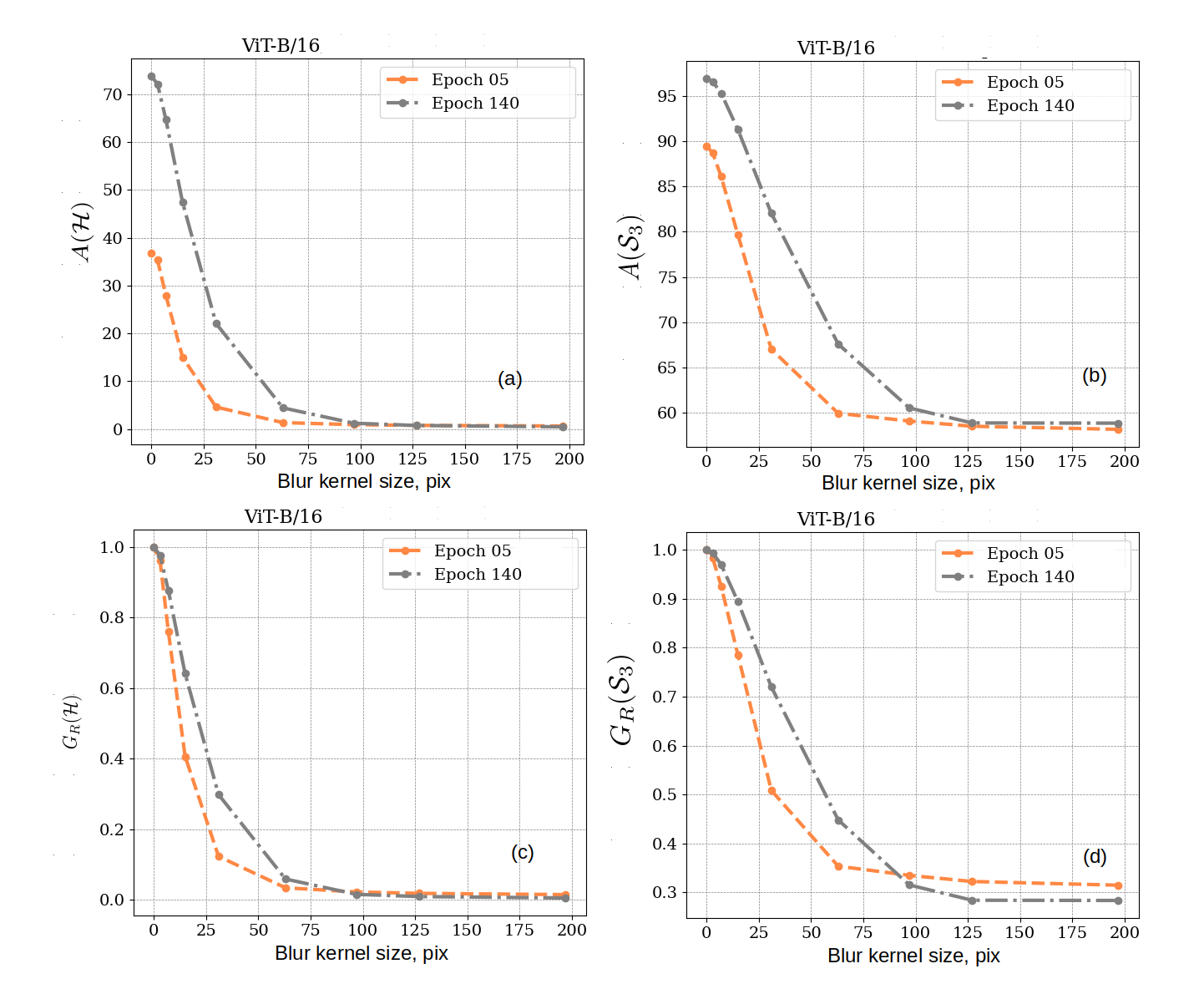}
  \caption{\label{fig_blur_vit}Accuracy (a,b) and relative accuracy gain
    (c,d) of ViT-B/16 in hyponym label space (a,c) and hypernym label space
    (b,d) with the respect to size of the blur kernel. $G_R$ is computed with
    respect to accuracy on undistorted images for the same checkpoint}
\end{figure}

As can be seen from the relative accuracy gain graph depicted in
\figurename~\ref{fig_blur_resnet} and \figurename~\ref{fig_blur_vit} there is no
tendency to depend on low frequencies more in the beginning of training than in
the end of training for both hypernyms and hyponyms.

\section{Complexity and infrastructure}
\label{appendix_complexity}

For experiments we used NVIDIA A6000 and RTX 3090 gpus.

Our framework consists of three components, each with its own computational
complexity:
\begin{enumerate}
\item \textbf{Greedy classifier model}. Involves single argmax computation over logits
  during forward pass. Mappings between labelspaces are efficiently managed
  using hash tables. Computational burden is negligible in practice.

\item \textbf{Manifold assessment}.
  - Complexity: $O(E L \cdot (N \log N \cdot p + K^2 d + c^2 K^2 ))$, where
  \begin{itemize}
  \item $E$ and $L$ are number of checkpoints and layers to consider respectively,
  \item $N=2K$ is number of data points (query and support sets),
  \item $c$ is number of classes,
  \item $p$ is feature size,
  \item $d$ is number of UMAP output dimensions.
  \end{itemize}

GPU acceleration is utilized for computing the distance matrix and mutual covers.
In our implementation, the experiments described in the paper required
approximately 40 GB of GPU memory.

\item \textbf{Neuronal collapse estimation}. Main computation burden comes from penultimate
  feature extraction. In our implementation number of forward passes: $(2 + M) S
  E$, where
  \begin{itemize}
   \item $S = 6 \times 10^5$ (size of balanced ImageNet),
   \item $M = 1$ (number of extra label spaces),
   \item $E = 350$ (checkpoints).
  \end{itemize}

\end{enumerate}

\end{document}